\documentclass[10pt,twocolumn,letterpaper]{article}

\usepackage[pagebackref=true,colorlinks,bookmarks=false]{hyperref}

\usepackage{cvpr}
\usepackage{times}
\usepackage{caption}
\usepackage{subcaption}
\usepackage{wrapfig}
\usepackage{amsmath}
\usepackage{amssymb}
\usepackage{amsthm}

\usepackage{algorithmic}
\usepackage{algorithm}

\usepackage[normalem]{ulem}

\usepackage{import}
\usepackage{tikz,tkz-base}
\usetikzlibrary{shapes,decorations,shadows}
\usetikzlibrary{decorations.pathmorphing}
\usetikzlibrary{decorations.shapes}
\usetikzlibrary{fadings}
\usetikzlibrary{patterns}
\usetikzlibrary{calc}
\usetikzlibrary{decorations.text}
\usetikzlibrary{decorations.footprints}
\usetikzlibrary{decorations.fractals}
\usetikzlibrary{shapes.gates.logic.IEC}
\usetikzlibrary{shapes.gates.logic.US}
\usetikzlibrary{fit,chains}
\usetikzlibrary{positioning}
\usepgflibrary{shapes}
\usetikzlibrary{scopes}
\usetikzlibrary{arrows}
\usetikzlibrary{backgrounds}

\newcommand{\DONE}[1]{}

\newcommand{\BB}{BB}

\usepackage{amsmath,amsfonts,bm,bbm,amssymb}
\usepackage{mathtools}
\DeclarePairedDelimiter{\ceil}{\lceil}{\rceil}
\newtheorem{theorem}{Theorem}
\newtheorem*{theorem*}{Theorem}
\newtheorem{lemma}{Lemma}
\newtheorem*{lemma*}{Lemma}

\usepackage{dsfont}
\usepackage{esvect}

\DeclareMathOperator*{\argmax}{arg\,max}

\newcommand{\SO}[1]{\ensuremath{\mathrm{SO(#1)}}}

\newcommand*\diff{\mathop{}\!\mathrm{d}}

\newcommand{\SDdim}[1]{\mathbb{S}^{#1}}

\def\cvprPaperID{1071} 

\newcommand{\tsum}{{\textstyle\sum}}

\ifcvprfinal\fi

\begin{document}

\title{Efficient Global Point Cloud Alignment using Bayesian Nonparametric Mixtures}

\author{
   Julian Straub$^*$ 
  \qquad Trevor Campbell\thanks{The first two
    authors contributed equally to this work.}
  \qquad Jonathan P. How
  \qquad John W. Fisher III\\
 Massachusetts Institute of Technology
}

\cvprfinalcopy

\maketitle

\begin{abstract}
  Point cloud alignment is a common problem in computer vision and
robotics, with applications ranging from 3D object recognition to reconstruction.
We propose a novel approach to the alignment problem that utilizes
Bayesian nonparametrics to describe the point cloud and surface
normal densities, and branch and bound (BB) optimization to recover the
relative transformation.
%
BB uses a novel, refinable, near-uniform tessellation of  
rotation space using 4D tetrahedra, leading
to more efficient optimization compared to the common axis-angle tessellation.
We provide objective function bounds for pruning given the proposed tessellation, and prove 
that BB converges to the optimum of the cost function along with providing its computational complexity. 
Finally, we empirically demonstrate the efficiency of 
the proposed approach as well as its robustness to
real-world conditions such as missing data and partial overlap.

\end{abstract}


\section{Introduction}

Point cloud alignment is a fundamental problem for
many applications in robotics~\cite{magnusson2007scan,henry2012rgb} and computer
vision~\cite{salvi2007review,newcombe2011kinectfusion,Whelan14ijrr}. 
Finding the global transformation is generally
hard: point-to-point correspondences typically do not exist, the
point clouds might only have partial overlap, and the underlying
objects themselves are often nonconvex, leading to 
a potentially large number of alignment local minima.
As such, popular local optimization techniques suffice only in
circumstances with small true relative transformations and large
overlap, such as in dense 3D incremental
mapping~\cite{henry2012rgb,newcombe2011kinectfusion,Whelan14ijrr}.
Solving the alignment problem for large unknown relative
transformations and small point cloud overlap calls for a global
approach. Example applications are the loop-closure problem in
SLAM~\cite{bosse2008map} and the model-based detection of objects in 3D
scenes~\cite{johnson1998surface}.

\begin{figure}
  \centering
  \includegraphics[width=.5\columnwidth]{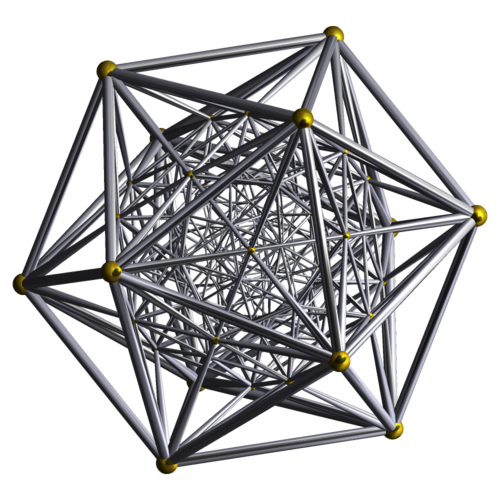}
  \caption{A 3D projection of the
  600-cell~\cite{StellaSoft}---a 4D object 
tessellating the space of rotations for the proposed branch and
bound approach to point cloud alignment.\label{fig:600cell}}
\end{figure}

Motivated by the observation that surface normal distributions are
translation invariant~\cite{horn1984extended} and straightforward to
compute~\cite{mitra2004estimating,straub2015rtmf},
we develop a two-stage branch and
bound~(\BB)~\cite{land1960automatic,lawler1966branch} optimization algorithm for point cloud
alignment.
%
We model the surface normal distribution of each point cloud as a
Dirichlet process (DP)~\cite{ferguson1973bayesian,Teh10_EML}
von-Mises-Fisher~(vMF)~\cite{fisher1995circularData}
mixture~\cite{straub2015dpvmf} (DP-vMF-MM). To find the optimal
rotation, we minimize the $L^2$ distance between the distributions over
the space of 3D rotations. We develop a novel refinable tessellation
consisting of 4D tetrahedra (see Fig.~\ref{fig:600cell}) which more uniformly
approximates rotation space and is more efficient than the common axis-angle
tessellation~\cite{li20073d,hartley2009global} during BB optimization.
Given the optimal rotation and modeling the two point distributions as
DP Gaussian mixtures~\cite{antoniak1974mixtures,chang13dpmm} (DP-GMM),
we obtain the optimal translation similarly via \BB{} over the space of
3D translations.
%
%
The use of mixture models circumvents discretization artifacts, while
still permitting efficient optimization.
In addition to algorithmic developments, we provide corresponding
theoretical bounds 
 on the convergence of both BB stages, linking the
quality of the derived rotation and translation estimates to the depth
of the search tree and thus the computation time of the algorithm. 
Experiments on real data corroborate the theory, and
demonstrate the accuracy and efficiency of BB as well as its robustness
to real-world conditions, such as partial overlap, high noise, and
large relative transformations.

\section{Related Work}

\noindent\textbf{Local Methods} \ \ 
There exists a variety of approaches for local point cloud
alignment~\cite{campbell2001survey,salvi2007review}. Iterative
closest point (ICP)~\cite{besl1992method}, the most common of these,
alternates between associating the points in both clouds
and updating the relative transformation estimate under those
associations. 
There are many variants of ICP~\cite{rusinkiewicz2001efficient}
differing in their choice of cost function, how correspondences are established,
and how the objective is optimized at each
iteration.
An alternative developed by
Magnusson~et~al.~\cite{magnusson2007scan} relies on the normal
distribution transform~(NDT)~\cite{biber2003normal}, which represents
the density of the scans as a structured GMM.
This approach has been shown to be more robust
than ICP in certain cases~\cite{magnusson2009evaluation}.  Approaches
that use correlation of kernel density estimates (KDE) for
alignment~\cite{tsin2004correlation} or GMMs~\cite{jian2011robust} use
a similar representation as the proposed approach. KDE-based methods
scale poorly with the number of points. In
contrast, we use mixture models inferred by
nonparametric clustering algorithms (DP-means~\cite{jordan2012dpmeans}
and DP-vMF-means~\cite{straub2015dpvmf}).  This allows adaptive
compression of the data, enabling the processing of large noisy point
clouds (see Sec.~\ref{sec:resultseval} for experiments with more than
$300$k points).  Straub et al.\@ propose two local rotational alignment
algorithms~\cite{straub2015dpvmf,straub2015rtmf} that, similarly to the
proposed approach, utilize surface normal distributions modeled as
vMF mixtures. Common to all local methods is the assumption of an
initialization close to the true transformation and significant overlap
between the two point clouds. If either of these assumptions are
violated, local methods become unreliable as they tend to get stuck in
suboptimal local
minima~\cite{rusinkiewicz2001efficient,salvi2007review,magnusson2009evaluation}.

\noindent\textbf{Global Methods} \ \ 
Global point cloud alignment algorithms make no prior assumptions about the relative
transformation or amount of overlap. For
those reasons global algorithms, such as the proposed one, are often
used to initialize local methods.
%
3D-surface-feature-based
algorithms~\cite{rusu2009fast,gelfand2005robust,johnson1998surface,aiger20084}
involve extracting local features,
obtaining matches between features in the two point clouds, and finally
estimating the relative pose using RANSAC~\cite{fischler1981ransac} or
other robust estimators~\cite{huber1981robust}. 
Though popular, feature-based algorithms are 
vulnerable to large fractions of incorrect feature matches, 
as well as repetitive scene elements and textures.
A second class of approaches, including the proposed approach, rely on
statistical properties of the two point clouds.
Makadia~et~al.~\cite{makadia2006fully} separate rotational and
translational alignment. Rotation is obtained by maximizing the convolution of the
peaks of the extended Gaussian images~(EGI)~\cite{horn1984extended} of
the two surface normal sets. This search is performed using
the spherical Fourier Transform~\cite{driscoll1994computing}. 
After rotational alignment, the translation is found similarly via the
fast Fourier Transform.
The use of histogram-based density estimates for the surface normal
and point distributions introduces discretization
artifacts. Additionally, the sole use of the peaks of the EGI makes the
method vulnerable to noise in the data.
For the alignment of 2D scans, Weiss~et~al.~\cite{weiss1994keeping} and
Bosse~et~al.~\cite{bosse2008map} follow a similar convolution-based
approach.
Early work by Li, Hartley and Kahl~\cite{li20073d,hartley2009global} on
BB for point cloud alignment used the
axis-angle (AA) representation of rotations. A drawback of this approach is
that a uniform AA tessellation does not lead to a uniform tessellation
in rotation space (see Sec.~\ref{sec:tessellationS3}). As we show in
Sec.~\ref{sec:resultseval}, this leads to less efficient BB search. 
Parra et al.~\cite{parra2014fast} propose improved bounds for
rotational alignment by reasoning carefully about the geometry of the
AA tessellation.  GoICP~\cite{yang2013go} nests BB over translations
inside BB over rotations and utilizes ICP internally to improve the BB
bounds. GOGMA~\cite{Campbell_2016_CVPR} uses a similar approach,
but replaces the objective with a convolution of GMMs. Both GoICP and GOGMA
involve BB over the joint 6-dimensional rotation and translation space; since the complexity 
of BB is exponential in the dimension, these methods are relatively computationally
expensive (see results Fig.~\ref{fig:apartment}). 



%
\section{The Point Cloud Alignment Problem}
Our approach to point cloud alignment relies on the fact that surface normal
distributions are invariant to translation~\cite{horn1984extended} and
easily computed~\cite{mitra2004estimating,straub2015rtmf}, allowing us
to isolate the effects of rotation. 
Thus we decompose the task of finding the relative transformation into first
finding the rotation using only the surface normal
distribution, and then obtaining the translation given
the optimal rotation.

Let a noisy sampling of a surface $S$ be described by the joint point
and surface normal density $p(x,n)$, where $x\in\mathbb{R}^3$ and
$n\in\SDdim{2}$.  
A sensor observes two independent samples
from this model: one from $p_1(x, n) = p(x, n)$, and one from $p_2(x,
n) = p(R^{\star T}(x-t^\star), R^{\star T} n)$ differing in an unknown
rotation $R^\star\in\SO3$ and translation $t^\star\in\mathbb{R}^3$.
Given these samples, we model the marginal point densities $\hat
p_1(x)$, $\hat p_2(x)$ using the posterior of a Dirichlet process
Gaussian mixture (DP-GMM)~\cite{antoniak1974mixtures}, and model the
marginal surface normal densities $\hat p_1(n)$, $\hat p_2(n)$ using
the posterior of a Dirichlet process von Mises-Fisher mixture
(DP-vMF-MM)~\cite{bangert2010using,straub2015dpvmf}.
Note that the formulation using DP mixture models admits
arbitrarily accurate estimates of a large class of noisy surface
densities (Theorem 2.2 in \cite{Devroye:1987:CDE:27672}).
Given the density estimates, we formulate the problem of finding
the relative transformation as
\begin{align} 
  \begin{aligned}\label{eq:alignment}
    \hat q &= \argmax_{q\in \SDdim{3}} \int_{\mathbb{S}^2} \hat p_1(n)\hat p_2(q\circ n)\mathrm{d}n \\
    \hat{t} &= \argmax_{t\in\mathbb{R}^3} \int_{\mathbb{R}^3} \hat p_1(x)\hat p_2(\hat q \circ x+t) \mathrm{d}x,
  \end{aligned}
\end{align} 
where we represent rotations using unit quaternions  
%
%
in $\SDdim{3}$, the 4D sphere~\cite{horn2001some},
and where $q\circ n$ denotes the rotation of a surface normal $n$ by a unit quaternion $q$.
Eq.~\eqref{eq:alignment} minimizes the $L_2$ metric via maximization of the
convolution, which has been shown to
be robust in practice~\cite{jian2011robust}.  This is a common approach
for Gaussian
MMs~\cite{tsin2004correlation,jian2011robust,Campbell_2016_CVPR} but to
our knowledge has not been explored for vMF-MMs, nor for Bayesian
nonparametric DP mixtures.  In fact, the use of DP mixtures is
critical, as it allows the automatic selection of a parsimonious, but
accurate, representation of the point cloud data. This improves upon
both kernel density estimates~\cite{tsin2004correlation}, which are
highly flexible but make optimizing Eq.~\eqref{eq:alignment}
intractable for large RGB-D datasets, and fixed-sized
GMMs~\cite{jian2011robust,Campbell_2016_CVPR}, which require heuristic
model selection and may not be rich enough to capture complex scene
geometry.
%
%
%
While exact posterior predictive DP-MM densities cannot be computed
tractably, excellent estimation algorithms are available, which we use
in this work~\cite{jordan2012dpmeans,straub2015dpvmf}. 

Both optimization problems in Eq.~\eqref{eq:alignment} are nonconcave
maximizations. Considering the geometry of the problem, we expect many
local maxima, rendering typical gradient-based methods ineffective.
This motivates the use of a global approach. We develop a 
two-step BB procedure~\cite{land1960automatic,lawler1966branch} that
first searches over $\mathbb{S}^3$ for the optimal rotation $\hat q$, and then over
$\mathbb{R}^3$ for the optimal translation $\hat t$. 
As BB may return multiple optimal
rotations (e.g.~if the scene has rotational symmetry) we estimate the optimal translation
under each of those rotations, and return the joint 
transformation with the highest translational cost lower bound. 
Note that while $\hat q, \hat t$ is not necessarily the optimal transformation
under rotation and translation \emph{jointly}, the decoupling of rotation
and translation we propose reduces the computational complexity
of \BB{} significantly. This is because the complexity  
scales exponentially in the search space dimension;
optimizing over two 3D spaces ($\mathbb{R}^3$ and $\SDdim{3}$) separately is significantly less costly 
than over the joint 6D space.
%

\BB{} requires three major components: (1) a
tessellation method for covering the optimization domain with
subsets (see Sec.~\ref{sec:tessellationS3} and
~\ref{sec:tessellationR3});  (2) a branch/refinement procedure for subdividing any subset
into smaller subsets (see Sec.~\ref{sec:tessellationS3} and
~\ref{sec:tessellationR3}); and (3) upper and lower bounds of the maximum objective on
each subset to be used for pruning (see Sec.~\ref{sec:vMFbounds} and \ref{sec:normalbounds}).  
\BB{} proceeds by bounding the optimal objective in each
subset, pruning those which cannot contain the maximum, subdividing the
best subset to refine the bounds, and iterating.
Note that in this work we select the node with the highest upper bound
for subdivision. More nuanced strategies have been developed and could
also be utilized~\cite{ibaraki1976theoretical,lawler1966branch}.
\section{vMF Mixture Rotational Alignment}

We model the distributions of surface normals $n$ as
von-Mises-Fisher~\cite{fisher1995circularData} mixture
models (vMF-MM) with means $\{\mu_{ik}\}_{k=1}^{K_i}$,
concentrations $\{\tau_{ik}\}_{k=1}^{K_i}$, and
positive weights $\{\pi_{ik}\}_{k=1}^{K_i}$, $\tsum_{k=1}^{K_i}\pi_{ik}
= 1$, for $i\in\{1, 2\}$, with density
\begin{align}
    \hat{p}_i(n) &= \textstyle\sum_{k=1}^{K_i}\pi_{ik} C_{ik}
    e^{\tau_{ik}\mu_{ik}^T n}\quad
    C_{ik} \triangleq\tfrac{\tau_{ik}}{4\pi \sinh(\tau_{ik})}.
\end{align}
While there are many techniques for inferring
vMF-MMs~\cite{banerjee2005clustering,dhillon2001concept,straub2015dpvmf},
we use a nonparametric method~\cite{straub2015dpvmf} that infers
an appropriate $K_i$ automatically.
%
The rotational alignment problem from Eq.~\eqref{eq:alignment} with this model becomes
\begin{align}
  \begin{aligned}\label{eq:rotAlignvMF}
    \max_{q\in \SDdim{3}}&\; 
    \tsum_{k, k'} \tfrac{D_{kk'}}{2\pi}\textstyle\int_{\mathbb{S}^2}
    e^{(\tau_{1k}\mu_{1k} + \tau_{2k'} q\circ
      \mu_{2k'})^Tn} \diff n \\
     D_{kk'} &\triangleq (2\pi)\pi_{1k}\pi_{2k'}C_{1k}C_{2k'}\,.
  \end{aligned}
\end{align}
We obtain the following objective function by 
noting that the integral is the normalization constant of a vMF 
density with concentration ${z_{kk'}(q) \triangleq
\|\tau_{1k}\mu_{1k} + \tau_{2k'} q\circ \mu_{2k'}\|}$:
%
\begin{align}
\begin{aligned}\label{eq:rotopt}
  \max_{q\in \SDdim{3}}\; &\tsum_{k, k'}D_{kk'}f(z_{kk'}(q)) \\
  \;\text{ where }\;&f(z) \triangleq 2 \sinh(z) z^{-1} = \left(e^{z} - e^{-z}\right) z^{-1}\,.
\end{aligned}
\end{align}




\subsection{Cover and Refinement of the Rotation Space
  $\mathbb{S}^3$\label{sec:tessellationS3}}

In this section, we develop a novel tessellation scheme for the space of rotations,
and show how to refine it in a way that guarantees convergence of \BB{} 
for rotational alignment. 
We follow a similar approach to the geodesic grid tessellation of a
sphere in 3D (i.e.~$\SDdim{2}$): as depicted in
Fig.~\subref{fig:tessellationS2}, starting from an icosahedron, each of
the $20$ triangular faces is subdivided into four triangles of equal
size.  Then the newly created triangle corners are normalized to unit
length, projecting them onto the unit sphere. 

\begin{figure}
  \setcounter{figure}{1}
  \renewcommand{\thesubfigure}{\arabic{figure}\alph{subfigure}}
  \centering
  \begin{subfigure}[b]{0.48\textwidth}
    \centering
    \begin{tikzpicture}
      \node (L0) at (0,0) {
        \includegraphics[height=0.24\columnwidth]{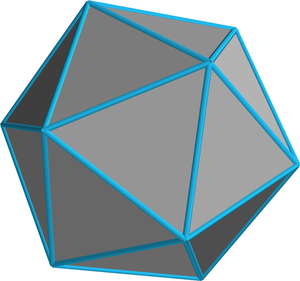}
      };
      \node[right = -0.5ex of L0] (L1) {
        \includegraphics[height=0.24\columnwidth]{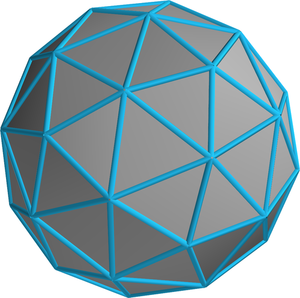}
      };
      \node[right = -0.5ex of L1] (L2) {
        \includegraphics[height=0.24\columnwidth]{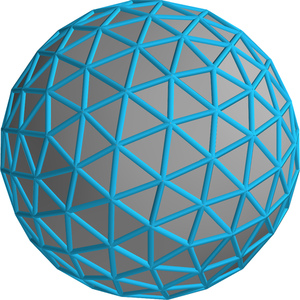}
      };
      \node[above = -0.5ex of L0] {Icosahedron};
      \node[above = -0.5ex of L1] {Subdiv. 1};
      \node[above = -0.5ex of L2] {Subdiv. 2};
      \node[below = -0.5ex of L0] (T0) {
        \includegraphics[height=0.24\columnwidth]{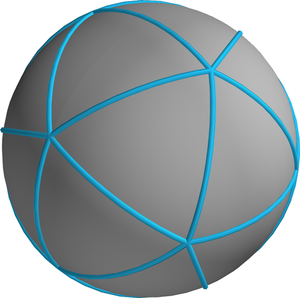}
      };
      \node[below = -0.5ex of L1] {
        \includegraphics[height=0.24\columnwidth]{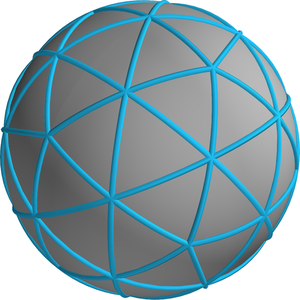}
      };
      \node[below = -0.5ex of L2] {
        \includegraphics[height=0.24\columnwidth]{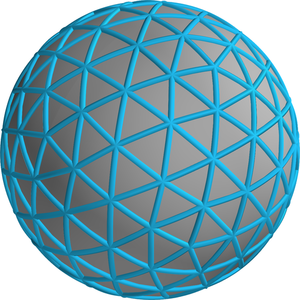}
      };
      \node (Q) at (0,-2.2) {$\mathcal{Q}$};
      \node[left = 1.0ex of L0,rotate=90,anchor=base ] {Triangles};
      \node[left = 1.0ex of T0,rotate=90,anchor=base ] {Tessellation};
    \end{tikzpicture}
    \caption{Tessellation of $\SDdim{2}$ via
      iterated triangle subdivision. The tessellation of $\SDdim{3}$ follows
      the same principles, but with 4D tetrahedra instead of 3D triangles. 
Note the uniformity of the tessellation.\label{fig:tessellationS2}}
\vspace{10pt}
  \end{subfigure} \quad
  \begin{subfigure}[b]{0.48\textwidth}
  \centering
  \begin{tikzpicture}
    \node (L0) at (0,0) {
      \includegraphics[height=0.24\columnwidth]{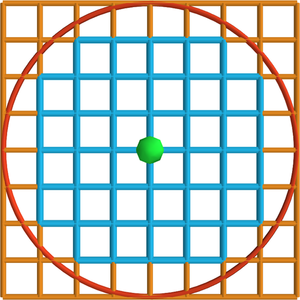}
    };
    \node[right = -0.5ex of L0] (L1) {
      \includegraphics[height=0.24\columnwidth]{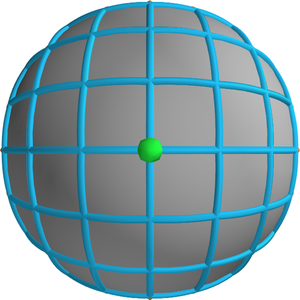}
    };
    \node[right = -0.5ex of L1] (L2) {
      \includegraphics[height=0.24\columnwidth]{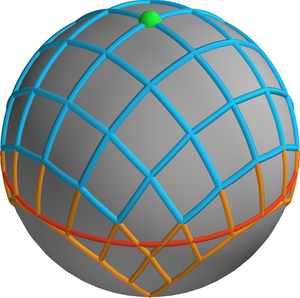}
    };
    \node[anchor=base,above = -0.5ex of L1] (Capt) {Top View};
    \node[anchor=base] at (Capt.base-|L0) {AA Space};
    \node[anchor=base] at (Capt.base-|L2) {Side View};
  \end{tikzpicture}

  \caption{Tessellation of $\SDdim{2}$ via uniform tessellation in the
    axis-angle (AA) space. 
The axis-angle tessellation
    of $\SDdim{3}$ follows the same principle and incurs similar
    distortion.
    Note 
    that orange tiles contain surface area on the lower
    half-sphere, so parts of the rotation space are covered
    twice, making \BB{} inefficient.
  \label{fig:tessellationAA}}
\end{subfigure}
\end{figure}


In four dimensions we instead start with the analogue of the
icosahedron, the \emph{600-cell}~\cite{coxeter1973regular} (shown in
Fig.~\ref{fig:600cell}), an object composed of 600 4D tetrahedra.  
We first generate its 120 vertices with the following algorithm~\cite[pp.~402--403]{coxeter1973regular}.
Let $\phi = \frac{1}{2}\left(1+\sqrt{5}\right)$. Then the (unnormalized) 120 vertices
of the 600-cell in 4D are
\begin{minipage}{\columnwidth}
\vspace{-4pt}
\begin{itemize}
\setlength{\itemsep}{-5pt}
\item even permutations of $\left[\pm\phi, \!\pm 1, \!\pm\phi^{-1}\!\!, 0\right]^T$ (96 vertices),
\item all permutations of $\left[\pm 2, 0, 0, 0\right]^T$ (8 vertices), and
\item all permutations of $\left[\pm 1, \pm 1, \pm 1, \pm 1\right]^T$ (16 vertices).
\end{itemize}
\vspace{-6pt}
\end{minipage}
We then scale the 120 vertices to each have unit norm, representing a 3D quaternion
rotation. Next, noting that the
angle between any two connected tetrahedra vertices is $36^\circ$,
we iterate over all $120 \choose 4$ possible choices of 4 vertices,
and only select those $600$ tetrahedra for which all pairwise angles are $36^\circ$.  
This collection of tetrahedra, which are ``flat'' in
4D analogous to triangles in 3D, comprises a 4D object which approximates
the 4D sphere, $\mathbb{S}^3$.
Then, since the set of all quaternion rotations may be represented by any hemisphere of $\SDdim{3}$ 
($q$ and $-q$ describe the same rotation),
we define the ``north'' vector to be $[0,\, 0,\, 0,\, 1]^T \in \SDdim{3}$, and only keep those
tetrahedra for which at least one vertex has angle $< 90^\circ$ to the north vector. 
This results in 330 tetrahedra that approximate the 4D upper hemisphere in $\mathbb{S}^3$, i.e.~the space of quaternion rotations.
Note that this construction procedure is the same for \emph{any} optimization
on $\mathbb{S}^3$, so it can be performed once and the result may be stored for
efficiency.

One major advantage of the proposed $\SDdim{3}$ tessellation is that it is
exactly uniform at the 0th level and approximately
uniform for deeper subdivision levels (Fig.~\subref{fig:tessellationS2} 
shows the analogous near-uniformity for $\SDdim{2}$). 
This generally tightens bounds employed by BB, leading to more
efficient optimization.  Another advantage is that this tessellation is
a near-exact covering of the upper hemisphere of $\SDdim{3}$. Only 7\%
of rotation space is covered twice, 
meaning that BB wastes little time with duplicate searching.
The widely employed AA-tessellation
scheme~\cite{li20073d,hartley2009global,parra2014fast,yang2013go}, in contrast, 
uniformly tessellates a cube enclosing the axis-angle space, a 3D
sphere with radius $\pi$, and maps that tessellation onto the rotation
space.
There are two major issues with the AA approach.
First, it covers 46\% of rotation space twice~\cite{li20073d,hartley2009global}
(see Fig.~\subref{fig:tessellationAA}).
Second, it does not lead to uniform
tessellation in rotation space. The reason for this is that the
Euclidean metric in AA space is a poor approximation of the distance on
the rotation manifold~\cite{li20073d}.  
Fig.~\subref{fig:tessellationAA} shows the AA
tessellation analog for $\SDdim{2}$, highlighting its significant non-uniformity.  We empirically find that the $\SDdim{3}$
tessellation leads to more efficient \BB{} optimization than the AA
tessellation (see results in Figs.~\ref{fig:bunnyFull} and
\ref{fig:bunny_045}).

We now discuss two properties of the proposed tessellation required by \BB{}:
1) that it is a \emph{cover} for the upper hemisphere of $\mathbb{S}^3$, guaranteeing
that \BB{} will search the whole space of rotations; and 2) that it is \emph{refinable},
so \BB{} can search promising subsets in increasingly more detail. 

\noindent\textbf{Cover} \ \  
Let the four vertices of a single tetrahedron from our approximation of
$\mathbb{S}^3$ be denoted $q_{j} \in\SDdim{3}$, $j\in\{1,\dots,4\}$. Then, stacking them horizontally into a matrix
$Q\in\mathbb{R}^{4\times 4}$, the projection $\mathcal{Q}$ of the
tetrahedron onto $\SDdim{3}$ is:
\begin{align}\label{eq:vertexlinearcombo}
  \mathcal{Q} &= \left\{q \in \mathbb{R}^4 : \|q\|=1, \, \, q=Q\alpha, \,  \,
  \alpha\in\mathbb{R}^4_+\right\}.
\end{align}
In other words, $\mathcal{Q}$ is the set of unit quaternions found by
extending the (flat in 4D) tetrahedron to the unit sphere using rays from the
origin. For $\SDdim{2}$, this is displayed in the second row of
Fig.~\subref{fig:tessellationS2}. 
The proposed set of 330 projected tetrahedra $\mathcal{Q}$ forms a cover of the upper hemisphere
of $\mathbb{S}^3$.

\begin{figure}
  \setcounter{figure}{2}
  \renewcommand{\thesubfigure}{\arabic{figure}\alph{subfigure}}
  \centering
  \begin{subfigure}[t]{0.48\textwidth}
    \centering
    \begin{tikzpicture}
      \node (L0) at (0,0) {
        \includegraphics[width=0.24\columnwidth]{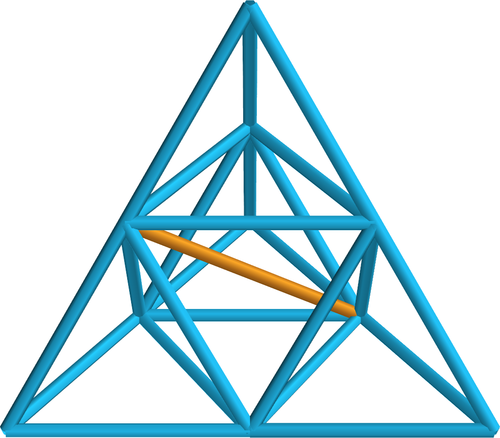}
      };
      \node[right = 1.5ex of L0] (L1) {
        \includegraphics[width=0.24\columnwidth]{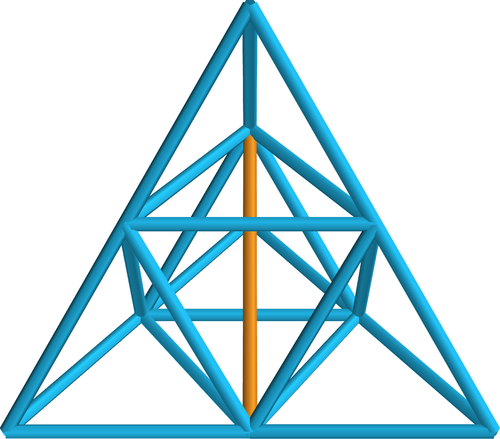}
      };
      \node[right = 1.5ex of L1] (L2) {
        \includegraphics[width=0.24\columnwidth]{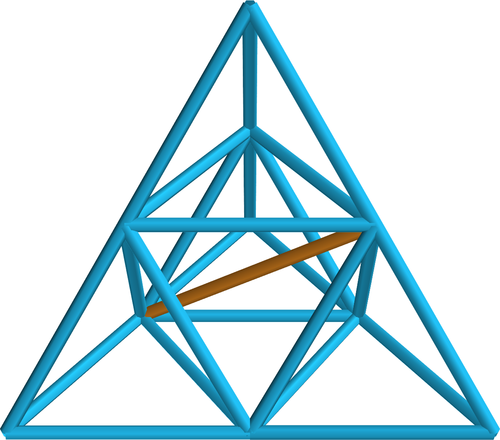}
      };
    \end{tikzpicture}
    \caption{The three subdivision patterns of a
      tetrahedron displayed in 3D. The internal orange edge is chosen
      to minimize distortion.
    \label{fig:tetrahedronSubdivision}}
   \vspace{10pt}
  \end{subfigure}\quad
  \begin{subfigure}[t]{0.48\textwidth}
    \centering
    \includegraphics[width=0.95\columnwidth]{{./figures/subdivisionVsMinAngle_ActualAndBound}.png}
    \caption{The bounds in Eq.~\eqref{eq:skew} compared to the true min \& max angles between
      tetrahedron vertices for
      increasing refinement level. \label{fig:minAngleBounds}}
  \end{subfigure}
\end{figure}

\noindent\textbf{Refinement} \ \ 
Next, we require a method of subdividing any $\mathcal{Q}$ 
in the cover. Similar to the triangle subdivision method for refining the
tessellation of $\SDdim{2}$, each 4D tetrahedron can be
subdivided into eight smaller tetrahedra~\cite{liu1996quality} as
depicted in Fig.~\subref{fig:tetrahedronSubdivision}.
The resulting six new vertices for the subdivided tetrahedra are 
scaled to unit length.
As we have the freedom to choose one of three internal edges
for subdivision, we choose the internal edge with the minimum
angle between its unit-norm vertices.
In other words, denoting $\xi_k$
for $k\in\{1, 2, 3\}$ to be the three internal dot products,
\begin{align}\label{eq:edgeselection}
  k^\star = \argmax_{k\in\{1, 2, 3\}} \,\, \xi_k.
\end{align}
This process forms the eight new subdivided cover elements $\mathcal{Q}$.
For example, if $q_i$, $i\in\{1,\dots,4\}$ are the vertices of $\mathcal{Q}$,
then one of the subdivisions (corresponding to one of the ``corner'' subtetrahedra 
in Fig.~\subref{fig:tetrahedronSubdivision}) of $\mathcal{Q}$
would have vertices
\begin{align}
q_1, \quad 
\frac{q_1+q_2}{\|q_1+q_2\|}, \quad
\frac{q_1+q_3}{\|q_1+q_3\|}, \quad \text{and} \quad
\frac{q_1+q_4}{\|q_1+q_4\|}.
\end{align}

Selecting the internal edge via Eq.~\eqref{eq:edgeselection}
is critical to our \BB{} convergence guarantee in Sec.~\ref{sec:rotconv}.
If Eq.~\eqref{eq:edgeselection} is not used, the individual
subsets $\mathcal{Q}$ can become highly skewed due to repeated
distortion from the unit-norm projection of the vertices, and refining 
$\mathcal{Q}$ does not necessarily correspond to shrinking the angular range 
of rotations it captures. Since we use Eq.~\eqref{eq:edgeselection},
however, Lemma~\ref{lem:skew} guarantees that subdividing 
$\mathcal{Q}$ shrinks its set of rotations appropriately:
\begin{lemma}\label{lem:skew}
Let $\gamma_N$ be the min dot product between vertices of 
any one $\mathcal{Q}$ at refinement level $N$. Then
\begin{align}\label{eq:skew}
  \tfrac{2\gamma_{N-1}}{1+\gamma_{N-1}} \leq \gamma_N, \qquad \text{where}\qquad \gamma_0 \triangleq \cos 36^\circ.
\end{align}
\end{lemma}
This result (proof in the supplement) shows that the tetrahedra
shrink and allow BB to improve its bounds during subdivision.
Figure~\subref{fig:minAngleBounds} demonstrates the tightness of
this bound, showing that $\cos^{-1}\gamma_N$ converges to 0 as $N\to\infty$. 
We conjecture that the \emph{max}
dot product $\Gamma_N$ satisfies a similar recursion,
 $\Gamma_N \leq \sqrt{(1+\Gamma_{N-1})/2}$, although this is not
required for our convergence analysis.
Fig.~\subref{fig:minAngleBounds} shows empirically that this 
matches the true max dot product,
but we leave the proof as an open problem.

\subsection{vMF Mixture Model Bounds\label{sec:vMFbounds}}
\BB{} requires both upper and lower bounds
on the maximum of the objective function within each projected
tetrahedron $\mathcal{Q}$, i.e.~we need $L$ and $U$ such that
\begin{align}
  L \leq \max_{q\in \mathcal{Q}} \textstyle\sum_{k, k'}D_{kk'}f(z_{kk'}(q))\leq
  U\,.
\end{align}
For the lower bound $L$, one can evaluate the
objective at \emph{any} point in $\mathcal{Q}$ (e.g.~its center). 
\begin{figure}
  \setcounter{figure}{3}
  \renewcommand{\thesubfigure}{\arabic{figure}\alph{subfigure}}
  \centering
  \subcaptionbox{The function $f(z)$ and its quadratic
upper bound, valid for $z \in \left[\ell_{kk'}, u_{kk'}\right]$
(here, $\ell_{kk'} = 1$ and $u_{kk'} = 4$).\label{fig:fbound}}[0.60\linewidth]{%
    \includegraphics[width=0.60\columnwidth]{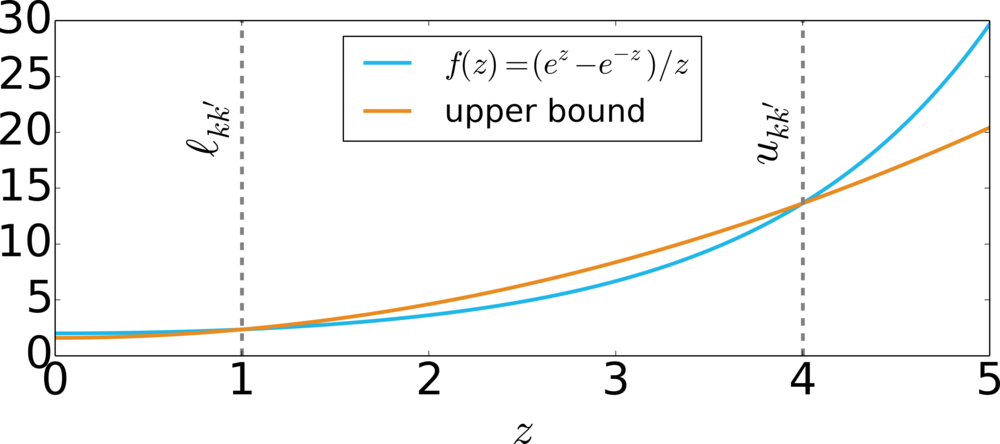}
  }
  \hspace{1ex}
  \subcaptionbox{Closest point (green) from a point $\mu$ (orange, Eq.~\ref{eq:ukkoptOnM}).\label{fig:closestpt} }[0.30\linewidth]{%
    \begin{tikzpicture}
      \node (L0) at (0,0) {
        \includegraphics[height=0.24\columnwidth]{{../figureClostestPointToM/closestPointToM_2_cropped}.png}
      };
      \node (m1) at (-0.6,0.9)  {$m_1$};
      \node (m2) at (0.7,0.6)   {$m_2$};
      \node (m3) at (0.4,-0.6)  {$m_3$};
      \node (m4) at (-0.7,-0.3) {$m_4$};
      \node (q) at (1.0,-0.2)   {$\mu$};
    \end{tikzpicture}
    }
  \setcounter{figure}{4}
\end{figure}

For the upper bound $U$, 
we use a quadratic upper bound on $f(z)$ (see Fig.~\subref{fig:fbound} and the supplement for details), noting that
$\ell_{kk'} \leq z_{kk'}(q) \leq u_{kk'}$ for all $q\in\mathcal{Q}$, where
\begin{align}\label{eq:LkkUkkdefn}
  \ell_{kk'}
  \triangleq
  \min_{q\in\mathcal{Q}}z_{kk'}(q) \quad \text{and}\quad
u_{kk'} \triangleq \max_{q\in\mathcal{Q}}z_{kk'}(q),
\end{align}
whose computation is discussed in Sec.~\ref{sec:likkuikk}.
This results in the upper bound $U$ where
\begin{align}
\begin{aligned}
  U &= \max_{q\in\mathcal{Q}} \, \, q^TAq + B\\
  A &\triangleq \tsum_{k,k'}2D_{kk'}\tau_{1k}\tau_{2k'}g_{kk'}\Xi_{kk'}
  \\
  B &\triangleq \tsum_{k,k'}D_{kk'}\left(
  (\tau_{1k}^2+\tau_{2k'}^2)g_{kk'}+h_{kk'}\right)\\
  g_{kk'} &\triangleq \tfrac{f(u_{kk'})-f(\ell_{kk'})}{u_{kk'}^2-\ell_{kk'}^2} \\
  h_{kk'}&\triangleq \tfrac{u_{kk'}^2f(\ell_{kk'})-\ell_{kk'}^2f(u_{kk'})}{u_{kk'}^2-\ell_{kk'}^2},\\
\end{aligned}
\end{align}
  and $\Xi_{kk'}\in\mathbb{R}^{4\times 4}$ is defined as the matrix for which ${\mu_{1k}^T(q\circ \mu_{2k'}) = q^T\Xi_{kk'} q}$
for any quaternion $q$ (see the supplement for details).
  Writing $q= Q\alpha$ as a linear combination of vertices of $\mathcal{Q}$
as in Eq.~\eqref{eq:vertexlinearcombo},
\begin{align}\label{eq:convexOptS3}
  \begin{aligned}
  U&=\max_{\alpha \in \mathbb{R}^4}\quad \alpha^TQ^TAQ\alpha +B \\
  &\quad\,\,\,\mathrm{s.t.}\quad \alpha^TQ^TQ\alpha = 1
  \,,\;\alpha \geq 0\,.
\end{aligned}
\end{align}
Since $\alpha\in\mathbb{R}^4$, and we have the constraint $\alpha \geq 0$,
we can search over all $\sum_{i=1}^4 {{4}\choose {i}}=15$ possible
combinations of components of $\alpha$ being zero or nonzero. Thus we solve the optimization
for $U_{\mathcal{I}}$ given
each possible subset $\mathcal{I}\subseteq \{1, 2, 3, 4\}$
of nonzero components of $\alpha$, and set
\begin{align}
U = B+\max_{\mathcal{I}\subseteq\{1, 2, 3, 4\}} U_{\mathcal{I}}.
\end{align}
For $U_{\mathcal{I}}$, we use a Lagrange multiplier for the equality constraint in Eq.~\eqref{eq:convexOptS3}
and set the derivative to 0, yielding a small generalized
eigenvalue problem of dimension $|\mathcal{I}|\leq 4$,
\begin{align}\label{eq:ges}
  \hspace{-.2cm}U_{\mathcal{I}}\!&=\! \max\! \left\{ \lambda : \exists v \geq 0, \, \left(Q^T\!\!AQ\right)_{\mathcal{I}}v = \lambda \left(Q^T\!Q\right)_{\mathcal{I}} v \right\}\!,
\end{align}
where $v$ is a $|\mathcal{I}|$-dimensional vector, and subscript $\mathcal{I}$ denotes the submatrix with rows and columns selected from $\mathcal{I}$.
The condition that all elements of $v$
are nonnegative in Eq.~\eqref{eq:ges} enforces that $\alpha \geq 0$ and
thus $\alpha$ corresponds to a solution $q$ that lies in $\mathcal{Q}$. 
Note that if $v$ is an eigenvector, so is $-v$.
If no $v$ satisfies $v\geq 0$, then we define $U_\mathcal{I} = -\infty$.

\subsection{Computing $\ell_{kk'}$ and $u_{kk'}$}\label{sec:likkuikk}
To find the upper bound $U$ in Eq.~\eqref{eq:convexOptS3}, we require the constants $\ell_{kk'}$ and $u_{kk'}$ 
for each pair of mixture components $k, k'$. Given their definitions in Eq.~\eqref{eq:LkkUkkdefn},
we have
\begin{align}\label{eq:ukklkkopt}
\hspace{-.3cm}{\small  \begin{aligned}
    u_{kk'} &= \sqrt{\tau_{1k}^2+\tau_{2k'}^2 + 2 \tau_{1k} \tau_{2k'}\max_{q\in\mathcal{Q}}\mu_{1k}^T(q\circ\mu_{2k'})}\,,\;\\
    \ell_{kk'} &= \sqrt{\tau_{1k}^2+\tau_{2k'}^2 -2 \tau_{1k} \tau_{2k'}\max_{q\in\mathcal{Q}}(-\mu_{1k})^T(q\circ\mu_{2k'})}\,.
  \end{aligned}
}
\end{align}
Since the inner optimization objective only depends on the rotation of $\mu_{2k'}$ by $q$,
we can reformulate the optimization as being over the set of 3D vectors $v\in\SDdim{2}$ such that $v = q\circ \mu_{2k'}$ for some $q\in\mathcal{Q}$.
Thus, finding $u_{kk'}$ and $\ell_{kk'}$ is equivalent to finding the closest and furthest unit vectors in 3D to $\mu_{1k}$ over the set of such vectors $v$, shown in Fig.~\subref{fig:closestpt}.
To solve this problem, let the vertices of $\mathcal{Q}$ be $q_i$, $i\in\{1, \dots, 4\}$,
and define the matrix 
$M \triangleq \left[m_1, \dots, m_4\right]\in\mathbb{R}^{3\times 4}$
where $m_i \triangleq q_i\circ \mu_{2k'}$. 
The inner optimization in Eq.~\eqref{eq:ukklkkopt} can be written as
(for $u_{kk'}$ set $\mu = \mu_{1k}$; for $\ell_{kk'}$ set $\mu=-\mu_{1k}$)
\begin{align}
\begin{aligned} \label{eq:ukkoptOnM}
J = \max_{\alpha \in \mathbb{R}^4} &\quad \mu^TM\alpha\\
\mathrm{s.t.}&\quad \alpha^TM^TM\alpha = 1 \quad \alpha \geq 0.
\end{aligned}
\end{align}
Showing that Eq.~\eqref{eq:ukkoptOnM}
is equivalent to solving the inner optimizations of Eq.~\eqref{eq:ukklkkopt}
is quite technical and is deferred to the supplement.
Again we search over all $\sum_{i=1}^3 {{4}\choose {i}}=14$ possible
combinations of components of $\alpha$ being zero or nonzero
(we do not check the $i=4$ case since in this case the matrix $M_{\mathcal{I}}$ 
below is rank-deficient).
We thus solve the optimization for $J_\mathcal{I}$
given each subset $\mathcal{I}\subseteq\{1, \dots, 4\}$, $|\mathcal{I}| \leq 3$ of nonzero components,
and set 
\begin{align}\label{eq:Jdefn}
J &= \max_{\mathcal{I}\subseteq\{1, 2, 3, 4\} \text{ s.t. } |\mathcal{I}|\leq 3} J_\mathcal{I}.
\end{align}
To solve for $J_\mathcal{I}$, we use
a Lagrange multiplier for the equality constraint,
and set derivatives to 0 to find that
\begin{align}
J_\mathcal{I} &= \sigma \sqrt{\mu^TM_\mathcal{I}\left(M_{\mathcal{I}}^TM_{\mathcal{I}}\right)^{-1}M_{\mathcal{I}}^T\mu}
\shortintertext{where}
\sigma &= \left\{\begin{array}{rl}
  1 & \left(M_{\mathcal{I}}^TM_{\mathcal{I}}\right)^{-1}M_{\mathcal{I}}^T\mu \geq 0\\
  -1 & \left(M_{\mathcal{I}}^TM_{\mathcal{I}}\right)^{-1}M_{\mathcal{I}}^T\mu \leq 0\\
  -\infty & \text{else}\, ,
\end{array}\right.
\end{align}
and $M_\mathcal{I}$ is the matrix constructed from the set of columns in $M$ corresponding to $\mathcal{I}$.
Note that $\sigma$ is also defined to be $\sigma = -\infty$ if $M_{\mathcal{I}}^TM_{\mathcal{I}}$ is not invertible.
After solving for the value of $J$ via Eq.~\eqref{eq:Jdefn}, we substitute it back into Eq.~\eqref{eq:ukklkkopt}
to obtain $u_{kk'}$ or $\ell_{kk'}$ as desired.

\subsection{Convergence Properties}\label{sec:rotconv}
We have now developed all the components necessary to
optimize Eq.~\eqref{eq:rotopt} via \BB{} on $\SDdim{3}$.
Theorem \ref{thm:epsconvergenceS3} (proof in the supplement)
provides a bound on the worst-case 
search tree depth $N$ to guarantee \BB{} terminates with 
rotational precision of $\epsilon$ degrees,
along with the overall computational complexity.
Note that the complexity of \BB{} is exponential in $N$,
but since $N$ is logarithmic in $\epsilon^{-2}$ 
 (by Theorem~\ref{thm:epsconvergenceS3},
Eq.~\eqref{eq:S3searchDepth} and $\cos x \simeq 1-x^2$ for $x\ll 1$), 
the complexity of \BB{} is polynomial in $\epsilon^{-1}$. 
 Recall from Sec.~\ref{sec:tessellationS3} that $\gamma_0$
for the 600-cell is $\gamma_0 \triangleq \cos 36^\circ$.

\begin{theorem}\label{thm:epsconvergenceS3}
  Suppose $\gamma_0$ is the initial maximum angle between vertices in
  the tetrahedra tessellation of $\SDdim{3}$, and let
  \begin{align}
N &\triangleq \max\left\{0,
  \ceil[\Big]{\log_2\tfrac{\gamma_0^{-1}-1}{\cos\left(\epsilon/2\right)^{-1}-1}} \right\}\,.
\label{eq:S3searchDepth}
  \end{align}
  Then at most $N$ refinements are required to 
  achieve an angular tolerance of $\epsilon$
 on $\mathbb{S}^2$, and \BB{} has complexity $O(\epsilon^{-6})$.
\end{theorem}

\section{Gaussian Mixture Translational Alignment}
In this section, we reuse notation for simplicity and to highlight parallels
between the translational and rotational alignment problems.
We model the density of points in the two point clouds as Gaussian mixture
models~(GMMs) with means $\{\mu_{ik}\}_{k=1}^{K_i}$, covariances $\{\Sigma_{ik}\}_{k=1}^{K_i}$,
and weights $\{\pi_{ik}\}_{k=1}^{K_i}$, $\sum_{k=1}^{K_i} \pi_{ik} = 1$, for $i\in\{1, 2\}$, 
with density
\begin{align}
    \hat{p}_i(x) = &\tsum_{k=1}^{K_i}\pi_{ik} \mathcal{N}(x; \mu_{ik}, \Sigma_{ik})\,.
\end{align}
GMMs can be inferred in a variety of
ways~\cite{jordan2012dpmeans,chang13dpmm}. 
Let $R^\star \in\SO{3}$ be the optimal rotation 
corresponding to $q^\star$ recovered using BB over $\SDdim{3}$.
Then defining
\begin{align}\label{eq:transzSm}\begin{aligned}
 m_{kk'} &\triangleq R^\star \mu_{2k'} - \mu_{1k} \,,\\
  S_{kk'} &\triangleq \Sigma_{1k} + R^\star \Sigma_{2k'} R^{\star T} \,,\\
z_{kk'}(t) &\triangleq
-\tfrac{1}{2}\left(t-m_{kk'}\right)^TS_{kk'}^{-1}\left(t-m_{kk'}\right)\,,
\end{aligned}
\end{align}
the translational optimization in Eq.~\eqref{eq:alignment} becomes:
\begin{align} 
\begin{aligned}\label{eq:translationalAlignmentGaussian}
  &\max_{t\in\mathbb{R}^3} \tsum_{k,k'} D_{kk'}f(z_{kk'}(t))\,\\
  &\text{where }f(z) \triangleq e^{z} \,,\;
  D_{kk'} \triangleq
\tfrac{\pi_{1k}\pi_{2k'}}{\sqrt{(2\pi)^{3}\left|S_{kk'}\right|}}\,.
\end{aligned}
\end{align}
This is again a nonconcave maximization, motivating the use of a global
approach.  Thus, we develop a second BB procedure on
$\mathbb{R}^3$ to find the optimal translation. 
%

\subsection{Cover and Refinement of $\mathbb{R}^3$\label{sec:tessellationR3}}
We tessellate the space of
translations, $\mathbb{R}^3$ with rectangular cells.
The initial tessellation is obtained by enclosing both point clouds
with a single rectangular bounding box with diagonal length
$\gamma_0$. 
For the refinement step, we choose to subdivide the
cell into eight equal-sized rectangular cells. Thus, the minimum $\gamma_N$ 
diagonal of the rectangular cells at refinement level $N$ 
possesses a straightforward shrinkage property similar to Eq.~\eqref{eq:skew},
\begin{align}\label{eq:transgammadefn}
  \tfrac{\gamma_{N-1}}{2} = \gamma_N.
\end{align}

\subsection{Gaussian Mixture Model Bounds\label{sec:normalbounds}}
As in the rotational problem, the translational \BB{} algorithm
requires lower and upper bounds on the
objective function in Eq.~\ref{eq:translationalAlignmentGaussian}:
\begin{align}
  L \leq \max_{t\in\mathcal{Q}} \tsum_{k,k'} D_{kk'}f(z_{kk'}(t)) \leq U \,.
\end{align}
For the lower bound $L$, one can evaluate the objective at \emph{any}
$t\in\mathcal{Q}$ (e.g.~its center). 
\begin{figure}
  \centering
  \includegraphics[width=0.8\columnwidth]{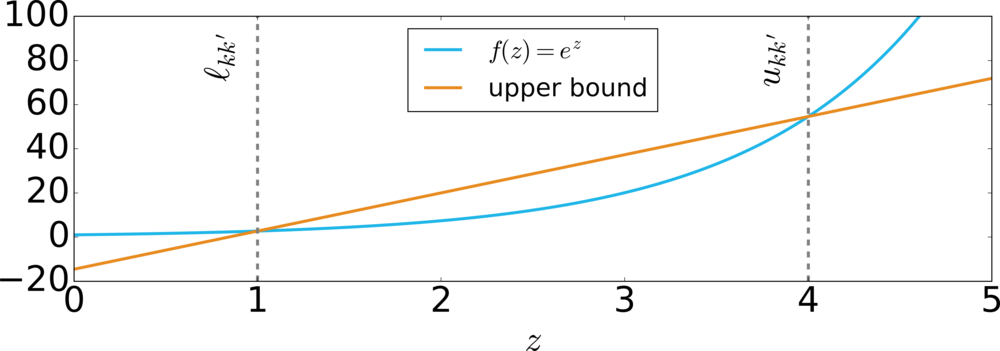}
  \caption{The function $f(z)$ and its linear
upper bound, valid for $z \in \left[\ell_{kk'}, u_{kk'}\right]$
(here, $\ell_{kk'} = 1$ and $u_{kk'} = 4$).\label{fig:tfbound}}
\end{figure}

For the upper bound $U$, 
we use a linear upper bound on $f(z)$ (see Fig.~\ref{fig:tfbound} and the supplement for details),
noting that $\ell_{kk'}\leq z_{kk'}(t) \leq u_{kk'}$ for all $q\in\mathcal{Q}$,
where
\begin{align}
   \ell_{kk'} \triangleq \min_{t\in\mathcal{Q}}z_{kk'}(t) 
\quad\text{and}\quad
u_{kk'} \triangleq \max_{t\in\mathcal{Q}}z_{kk'}(t)
\,,
\end{align}
whose computation is discussed in Section \ref{sec:tlikkuikk}.
This results in the upper bound $U$, where
\begin{align}
  \begin{aligned} \label{eq:upperConvNormalDetails}
U &\triangleq \max_{t\in\mathcal{Q}} t^TAt + B^Tt + C \\
  A &\triangleq -\tfrac{1}{2}\tsum_{k,k'}D_{kk'}g_{kk'}S_{kk'}^{-1}\\
  B &\triangleq \tsum_{k,k'}D_{kk'} g_{kk'}S_{kk'}^{-1}m_{kk'}\\
  C &\triangleq
  \tsum_{kk'}D_{kk'}\left(h_{kk'}-\tfrac{1}{2}g_{kk'}m_{kk'}^TS_{kk'}^{-1}m_{kk'}\right)\\
  g_{kk'} &\triangleq \tfrac{f(u_{kk'})-f(\ell_{kk'})}{u_{kk'}-\ell_{kk'}} \\
  h_{kk'} &\triangleq \tfrac{u_{kk'}f(\ell_{kk'})-\ell_{kk'}f(u_{kk'})}{u_{kk'}-\ell_{kk'}}\,.
\end{aligned}
\end{align}
%
%
This is a concave quadratic maximization over a rectangular cell $\mathcal{Q}$. Thus,
we obtain $U$ as the maximum over all local optima in the
interior, faces, edges, and vertices of $\mathcal{Q}$.


%
%
\subsection{Computing $\ell_{kk'}$ and $u_{kk'}$}\label{sec:tlikkuikk}
Using the form of $z_{kk}(t)$ in Eq.~\eqref{eq:transzSm}, 
we have that
\begin{align}
\hspace{-.3cm}
\begin{aligned}
\ell_{kk'} / u_{kk'} &= \min_{t\in\mathcal{Q}} / \max_{t\in\mathcal{Q}} \quad t^TAt + B^Tt+C\\
A \triangleq-\tfrac{1}{2}S^{-1}_{kk'}\,,\;
B &\triangleq -2 A m_{kk'}\,,\;
  C \triangleq -\tfrac{1}{2}m_{kk'}^T B.
\end{aligned}
\end{align}
Because of the concavity of the objective,
$u_{kk'}$ can be obtained with the exact same algorithm
as used to solve Eq.~\ref{eq:upperConvNormalDetails}.
$\ell_{kk'}$ can be obtained by checking the vertices of $\mathcal{Q}$,
as the minimum of a concave function over a rectangular cell must occur at one of its vertices.

\subsection{Convergence Properties}
We now have all the components necessary to
optimize Eq.~\eqref{eq:translationalAlignmentGaussian} via \BB{} on
$\mathbb{R}^3$. As in the rotational alignment case, we provide
a characterization (Theorem~\ref{thm:epsconvergenceR3}, proof in the supplement) of the maximum refinement depth  $N$
required for a desired translational precision $\epsilon$,
along with the complexity of the algorithm.
Note that while the complexity of \BB{} is exponential in $N$,
$N$ is logarithmic in $\epsilon^{-1}$ (Theorem~\ref{thm:epsconvergenceR3}),
so \BB{} has polynomial complexity in $\epsilon^{-1}$. 
\begin{theorem}\label{thm:epsconvergenceR3}
  Suppose $\gamma_0$ is the initial diagonal length of the translation cell
  in $\mathbb{R}^3$, and let
  \begin{align}
      N &\triangleq \max\left\{0, \ceil[\Big]{\log_2\tfrac{\gamma_0}{\epsilon}} \right\} \,.
  \end{align} 
  Then at most $N$ refinements are required to achieve 
a translational tolerance of $\epsilon$,
  and \BB{} has complexity $O(\epsilon^{-3})$.
\end{theorem}

%
\section{Results and Evaluation\label{sec:resultseval}}
We evaluate \BB{} (both with and without final local refinement~\cite{chen1991object}) on 
four datasets~\cite{curless1996volumetric,turk1994zippered,Pomerleau:2012}
compared to three global methods: an FT-based method~\cite{makadia2006fully}, GoICP~\cite{yang2013go} 
($20\%$ trimming), and GOGMA~\cite{Campbell_2016_CVPR}.
To generate the vMF-MMs and GMMs for \BB{},
we cluster the data with DP-vMF-means~\cite{straub2015dpvmf} and
DP-means~\cite{jordan2012dpmeans}, and fit
maximum likelihood MMs to the clustered data.
To account for nonuniform point densities due to the sensing
process, we weight each point's contribution to the MMs by its surface
area, estimated by the disc of radius equal to the fifth nearest neighbor distance.
We use kNN+PCA~\cite{meshlab,libpcl} to extract surface normals.
To improve the robustness of BB, it is run three times on each
problem with 
scale values $\lambda_n \in \{45^\circ, 65^\circ,
80^\circ\}$ in DP-vMF-means (included in the timing results).
The scale $\lambda_x$ for DP-means is
manually selected to yield around $50$
mixture components. 
Using Theorems~\ref{thm:epsconvergenceS3}~and~\ref{thm:epsconvergenceR3}, we
terminate rotational BB at $N=11$ and translational BB at $N=10$ for
a rotational accuracy of $1^\circ$ and a translational accuracy
of $\tfrac{\gamma_0}{1024}$, where $\gamma_0$ is defined in Eq.~\eqref{eq:transgammadefn}. 
All timing results include algorithm-specific
preprocessing of the data. We used a 3GHz core i7
CPU and a GeForce GTX 780 GPU. While
clustering via DP-means and DP-vMF-means uses the GPU, we only
use parallel CPU threads for the eight \BB{} bound evaluations after each branch step.  

\begin{figure}
  \centering
  \begin{tikzpicture}
    \node (1) at (0,0) {
      \includegraphics[width=0.40\linewidth]{{./figuresSmall/bb_iteration_stats_S3_bound_nodes_cropped}.png}
    };
    \node[right = 0ex of 1.north east,anchor=north west] (2)  {
      \includegraphics[width=0.40\linewidth]{{./figuresSmall/bb_iteration_stats_S3_area_time_cropped}.png}
    };

    \node[right = 0ex of 2.north east,anchor=north west] (before) {
      \includegraphics[width=0.08\linewidth]{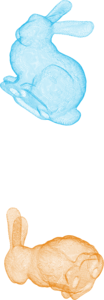}
    };
    \node[below = 1.5ex of before] (BB)  {
      \includegraphics[width=0.08\linewidth]{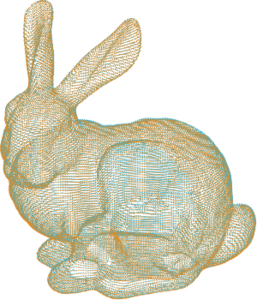}
    };
    \draw[->,line width=0.2mm] (before) to (BB);
  \end{tikzpicture}
  \caption{BB alignment of the full Stanford Bunny.\label{fig:bunnyFull}}
\end{figure}

%

\begin{figure}
  \centering
  \begin{tikzpicture}
    \node (1) at (0,0) {
      \includegraphics[width=0.40\linewidth]{{./figuresSmall/bb_iteration_stats_S3_bound_nodes_cropped}.png}
    };
    \node[right = 0ex of 1.north east,anchor=north west] (2)  {
      \includegraphics[width=0.40\linewidth]{{./figuresSmall/bb_iteration_stats_S3_area_time_cropped}.png}
    };

    \node[right = 0ex of 2.north east,anchor=north west] (before) {
      \includegraphics[width=0.08\linewidth]{{./figuresSmall/before_1_cropped}.png}
    };
    \node[below = 1.5ex of before] (BB)  {
      \includegraphics[width=0.08\linewidth]{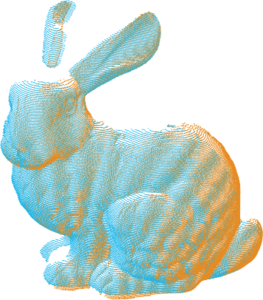}
    };
    \draw[->,line width=0.2mm] (before) to (BB);
  \end{tikzpicture}
%
  \caption{Alignment of partial scans of the Stanford
  Bunny.\label{fig:bunny_045}}
\end{figure}

%

\noindent\textbf{Stanford Bunny~\cite{turk1994zippered}} \ \ 
Independent of the tessellation strategy,
\BB{} perfectly aligns the Stanford Bunny with a
randomly transformed version of itself, as shown in Fig.~\ref{fig:bunnyFull}.
The results of aligning two partial scans of the Stanford Bunny with
relative viewpoint difference $45^\circ$ are shown in
Fig.~\ref{fig:bunny_045}. 
\BB{}'s initial alignment is close enough to allow ICP to
converge to a perfect alignment. 
%
The proposed approach leads to a faster reduction in the
bound gap, faster exploration, and a smaller number of active nodes, while reducing the computation time per iteration
by an order of magnitude vs the AA tessellation.
This shows conclusively that the proposed
tessellation leads to more efficient BB optimization. 
Note that the AA tessellation starts
at $146\%$ unexplored space because it covers the
rotation space more
than once as discussed in Sec.~\ref{sec:tessellationS3}.
In both cases BB finds the optimal translation within
$200$ iterations.

\begin{figure}
  \centering
  \subcaptionbox*{initial\label{fig:happB_init}}[0.155\linewidth]{%
    \includegraphics[height=0.09\textheight]{{./figuresSmall/before_1_cropped}.png}
  }
  \subcaptionbox*{BB\label{fig:happB_BB}}[0.155\linewidth]{%
    \includegraphics[height=0.10\textheight]{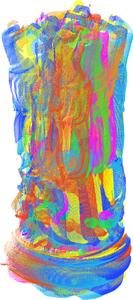}
  }
  \subcaptionbox*{BB+ICP\label{fig:happB_BB_ICP}}[0.155\linewidth]{%
    \includegraphics[height=0.10\textheight]{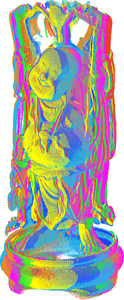}
  }
  \subcaptionbox*{GoICP\label{fig:happB_GoICP}}[0.155\linewidth]{%
    \includegraphics[height=0.10\textheight]{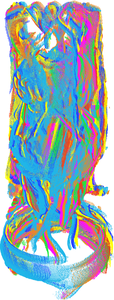}
  }
  \subcaptionbox*{GOGMA\label{fig:happB_GOGMA}}[0.155\linewidth]{%
    \includegraphics[height=0.10\textheight]{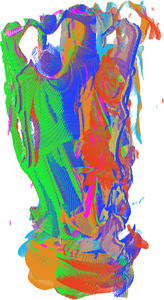}
  }
  \subcaptionbox*{FT\label{fig:happB_FT}}[0.155\linewidth]{%
    \includegraphics[height=0.09\textheight]{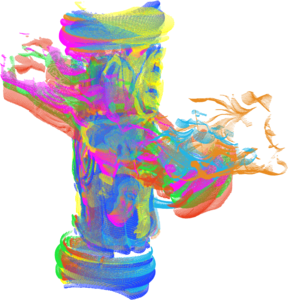}
  }
%
  \caption{Alignment of partial scans of Happy Buddha.
\label{fig:happyB}}
\end{figure}

\noindent\textbf{Happy Buddha~\cite{curless1996volumetric}} \ \ 
This dataset consists of 15 scans taken at $24^\circ$
rotational increments about the vertical axis of a statue. This
dataset is challenging, as the scans contain few overlapping
points, and the surface normal distributions are anisotropic.
 We perform pairwise alignment of consecutive scans, and render the aligned scans
together in one coordinate system (Fig.~\ref{fig:happyB}). 
The only successful alignment is produced by BB+ICP. 
This shows the advantage of using surface normals for rotational alignment.
Other methods using points (GoICP) or GMMs (GOGMA)
have difficulty dealing with ambiguities due to the ``flatness'' of the scans.

\begin{figure}
  \centering
  \includegraphics[height=0.13\textheight]{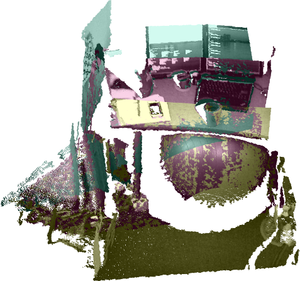}
  \includegraphics[height=0.13\textheight]{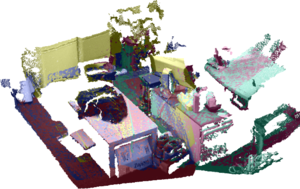}
  \caption{Correct alignment of five noisy, incomplete, and partially
  overlapping RGB-D point clouds of cluttered indoor scenes using
BB+ICP. Colors indicate different scans.\label{fig:rgbd}}
\end{figure}

\noindent\textbf{Office Scan} \ \ 
Figure~\ref{fig:rgbd} demonstrates that BB+ICP finds accurate registrations on
noisy, incomplete, cluttered and irregular point clouds as long as good
surface normal estimates are available. This demonstrates the potential
use of BB+ICP for loop closure detection.

\begin{figure}
  \includegraphics[width=0.48\linewidth]{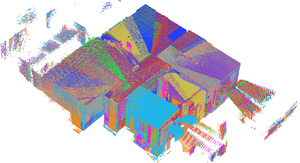}
  \includegraphics[width=0.48\linewidth]{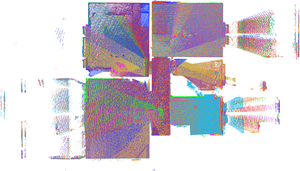}
  \caption{Apartment dataset~\cite{Pomerleau:2012} aligned using BB+ICP\label{fig:apartment}}
\end{figure}

\noindent\textbf{Apartment Dataset~\cite{Pomerleau:2012}} \ \ 
This dataset consists of $44$ LiDAR scans with an average overlap of $84\%$.
Figure~\ref{fig:apartment} shows the BB+ICP aligned scans of the dataset.
Table~\ref{tab:apartment} compares the accuracy 
and inlier percentages defined by (C)oarse ($2$m; $10^\circ$),
(M)edium ($1$m; $5^\circ$) and (F)ine ($0.5$m; $2.5^\circ$) thresholds for all algorithms.
For GoICP, we used $100$ scan points and an accuracy threshold of $0.01$. 
We used the scale parameter of $\lambda_x=1.3$m for GMM computations in both GOGMA and BB.

\begin{table}
  \scalebox{0.79}{
  \setlength{\tabcolsep}{3pt}
  \begin{tabular}{lrrrrrrrrrr}
     Method                     & $[*]_\lambda$   & $[*]_\lambda\scriptscriptstyle+$ 
                                & $[*]^M$   & $[*]^M\!\!\scriptscriptstyle+$ 
                                & $[*]_\lambda^M$ & $[*]_\lambda^M\!\!\scriptscriptstyle+$ 
                                & \cite{Campbell_2016_CVPR} & \cite{Campbell_2016_CVPR}$\scriptscriptstyle+$  & \cite{yang2013go} & \cite{makadia2006fully} \\\hline
    Rot [$^\circ\hspace{-1pt}$] &     28.6      &  26.9    &    5.52 &   1.61    &   3.77     &  \textbf{   1.36 }   &  7.14  &   5.14 & 24.2    & 30.0 \\
    Tran [\small{m}]            &     0.48      &  0.43    &    0.12 &   0.04    &   0.08     &  \textbf{   0.03 }   &  0.22  &   0.09 & 0.46    & 0.65 \\\hline
    Inl \% C                    &     79.6      &  81.8    &   90.9  &  95.5     &  93.2      &  \textbf{  97.7 }    & 97.5   &  97.5 & 47.7    & 29.5 \\  
    Inl \% M                    &     75.0      &  81.8    &   79.6  &  95.5     &  86.4      &  \textbf{  97.7 }    & 85.0   &  97.5 & 34.1    & 18.2 \\ 
    Inl \% F                    &     54.6      &  81.8    &   36.4  &  95.5     &  61.4      &  \textbf{  97.7 }    & 47.5   &  97.5 & 13.6    & 2.27 \\ \hline
    Time [s]                    & \textbf{32.6} &  50.0    &   38.4  &  57.3     & 140      &           156      &         405& 675 & 62.0    & 470
  \end{tabular}
}
\caption{Apartment~\cite{Pomerleau:2012} results using BB~[$*$],
GOGMA~\cite{Campbell_2016_CVPR}, GoICP~\cite{yang2013go}, and
FT~\cite{makadia2006fully}.
We denote search over rotational scale via $_\lambda$, search over MW ambiguities
with $^M$ and local refinement with $+$. We report rotational (Rot),
translational (Tran), timing, and inlier (Inl) percentages for (C)oarse,
(M)edium and (F)ine alignment (as defined in the text).
\label{tab:apartment}}
\end{table}

\begin{figure}
  \begin{tikzpicture}
    \node (dR) {
      \includegraphics[width=0.321\linewidth]{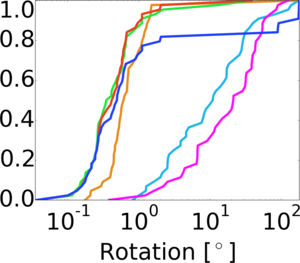}
    };
    \node[right = -1ex of dR] (dT) {
      \includegraphics[width=0.31\linewidth]{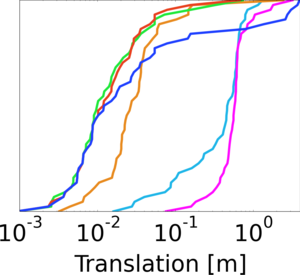}
    };
    \node[right = -1ex of dT] (dt) {
      \includegraphics[width=0.31\linewidth]{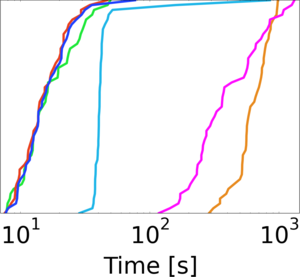}
    };
    \node[above = -0ex of dT,xshift=-0ex] (legend) {
      \includegraphics[width=0.95\linewidth]{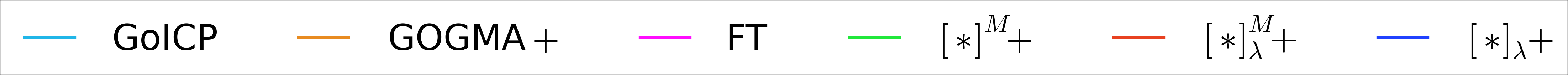}
    };
  \end{tikzpicture}
  \caption{Cumulative density functions of rotational error, translational error, and runtime. \label{fig:CDF}}
\end{figure}

Man-made environments such as this dataset exhibit ``Manhattan World'' (MW)
symmetry in their surface normal
distributions~\cite{straub2015rtmf,straub2014mmf}. We thus
 transform the rotation obtained via rotational BB by all $24$ 
MW rotations, and search over all using translational BB. 
Note that doing this is straightforward in the proposed decoupled BB
approach, as opposed to a joint approach, e.g.~GoICP and GOGMA.

Table~\ref{tab:apartment} and Fig.~\ref{fig:CDF} show that
BB with searching over both scale and MW rotations leads to the best accuracy among 
all algorithms, with a 3x speedup over the $2^{\text{nd}}$
best method, GOGMA (which uses a GPU).
From the inlier percentages it is clear that FT and GoICP do not perform well.
The CDFs in Fig.~\ref{fig:CDF} show that accounting for MW symmetry (red, green)
is important; ignoring it (blue) causes scans to be flipped by $90^\circ$/$180^\circ$,
affecting the mean error strongly. 

\section{Conclusion}
We introduced a BB approach to global point cloud
alignment with convergence guarantees, based
on a Bayesian nonparametric point cloud representation and 
a novel tessellation of rotation space.
The method decouples translation and rotation
via the use of surface normals, making it more
efficient than previous joint approaches.
Experiments demonstrate the robustness of
the method to noisy real world data, partial overlap, and
angular viewpoint differences.
We expect that the proposed tessellation of $\SDdim{3}$ will be useful in
other rotational \BB{} algorithms.
All code is available at~{\footnotesize\url{http://people.csail.mit.edu/jstraub/}}.

{\small
\bibliographystyle{ieee}

\begin{thebibliography}{}\itemsep=-1pt

\end{thebibliography}


\begin{thebibliography}{10}\itemsep=-1pt

\bibitem{aiger20084}
D.~Aiger, N.~J. Mitra, and D.~Cohen-Or.
\newblock 4-points congruent sets for robust pairwise surface registration.
\newblock In {\em ACM TOG}, volume~27, page~85,
  2008.

\bibitem{antoniak1974mixtures}
C.~Antoniak.
\newblock Mixtures of {Dirichlet} processes with applications to {Bayesian}
  nonparametric problems.
\newblock {\em The Annals of Statistics}, 1152--1174, 1974.

\bibitem{banerjee2005clustering}
A.~Banerjee, I.~S. Dhillon, J.~Ghosh, S.~Sra, and G.~Ridgeway.
\newblock Clustering on the unit hypersphere using {von Mises-Fisher}
  distributions.
\newblock {\em JMLR}, 6(9), 2005.

\bibitem{bangert2010using}
M.~Bangert, P.~Hennig, and U.~Oelfke.
\newblock Using an infinite {von Mises-Fisher} mixture model to cluster
  treatment beam directions in external radiation therapy.
\newblock In {\em {ICMLA}}, 2010.

\bibitem{besl1992method}
P.~J. Besl and N.~D. McKay.
\newblock A method for registration of 3-D shapes.
\newblock {\em TPAMI}, 14(2):239--256, 1992.

\bibitem{biber2003normal}
P.~Biber and W.~Stra{\ss}er.
\newblock The normal distributions transform: A new approach to laser scan
  matching.
\newblock In {\em IROS}, 2003.

\bibitem{bosse2008map}
M.~Bosse and R.~Zlot.
\newblock Map matching and data association for large-scale two-dimensional
  laser scan-based SLAM.
\newblock {\em IJRR}, 27(6):667--691, 2008.

\bibitem{Campbell_2016_CVPR}
D.~Campbell and L.~Petersson.
\newblock Gogma: Globally-optimal gaussian mixture alignment.
\newblock In {\em CVPR}, June 2016.

\bibitem{campbell2001survey}
R.~J. Campbell and P.~J. Flynn.
\newblock A survey of free-form object representation and recognition
  techniques.
\newblock {\em Computer Vision and Image Understanding}, 81(2):166--210, 2001.

\bibitem{chang13dpmm}
J.~Chang and J.~W. Fisher~III.
\newblock Parallel sampling of DP mixture models using sub-clusters splits.
\newblock In {\em NIPS}, 2013.

\bibitem{chen1991object}
Y.~Chen and G.~Medioni.
\newblock Object modeling by registration of multiple range images.
\newblock In {\em ICRA}, 1991.

\bibitem{coxeter1973regular}
H.~S.~M. Coxeter.
\newblock {\em Regular polytopes}.
\newblock Courier Corporation, 1973.

\bibitem{curless1996volumetric}
B.~Curless and M.~Levoy.
\newblock A volumetric method for building complex models from range images.
\newblock In {\em SIGGRAPH}, 1996.

\bibitem{Devroye:1987:CDE:27672}
L.~Devroye.
\newblock {\em A Course in Density Estimation}.
\newblock Birkhauser Boston Inc., 1987.

\bibitem{dhillon2001concept}
I.~S. Dhillon and D.~S. Modha.
\newblock Concept decompositions for large sparse text data using clustering.
\newblock {\em Machine Learning}, 42(1-2):143--175, 2001.

\bibitem{driscoll1994computing}
J.~R. Driscoll and D.~M. Healy.
\newblock Computing Fourier transforms and convolutions on the 2-sphere.
\newblock {\em Advances in Applied Mathematics}, 15(2):202--250, 1994.

\bibitem{ferguson1973bayesian}
T.~Ferguson.
\newblock A {Bayesian} analysis of some nonparametric problems.
\newblock {\em The Annals of Statistics}, 209--230, 1973.

\bibitem{fischler1981ransac}
M.~Fischler and R.~Bolles.
\newblock Random sample consensus: a paradigm for model fitting with
  applications to image analysis and automated cartography.
\newblock {\em Communications of the ACM}, 24(6):381--395, 1981.

\bibitem{fisher1995circularData}
N.~I. Fisher.
\newblock {\em Statistical Analysis of Circular Data}.
\newblock Cambridge University Press, 1995.

\bibitem{gelfand2005robust}
N.~Gelfand, N.~J. Mitra, L.~J. Guibas, and H.~Pottmann.
\newblock Robust global registration.
\newblock In {\em Symposium on Geometry Processing}, volume~2, page~5, 2005.

\bibitem{hartley2009global}
R.~I. Hartley and F.~Kahl.
\newblock Global optimization through rotation space search.
\newblock {\em IJCV}, 82(1):64--79, 2009.

\bibitem{henry2012rgb}
P.~Henry, M.~Krainin, E.~Herbst, X.~Ren, and D.~Fox.
\newblock RGB-D mapping: Using Kinect-style depth cameras for dense 3D modeling
  of indoor environments.
\newblock {\em IJRR}, 31(5):647--663, 2012.

\bibitem{horn2001some}
B.~K. Horn.
\newblock Some notes on unit quaternions and rotation.
\newblock 2001.

\bibitem{horn1984extended}
B.~K.~P. Horn.
\newblock Extended {Gaussian} images.
\newblock {\em Proceedings of the IEEE}, 72(12):1671--1686, 1984.

\bibitem{huber1981robust}
P.~J. Huber.
\newblock {\em Robust statistics}.
\newblock Springer, 1981.

\bibitem{ibaraki1976theoretical}
T.~Ibaraki.
\newblock Theoretical comparisons of search strategies in branch-and-bound
  algorithms.
\newblock {\em IJCIS}, 5(4):315--344, 1976.

\bibitem{jian2011robust}
B.~Jian and B.~C. Vemuri.
\newblock Robust point set registration using gaussian mixture models.
\newblock {\em PAMI}, 33(8):1633--1645, 2011.

\bibitem{johnson1998surface}
A.~E. Johnson and M.~Hebert.
\newblock Surface matching for object recognition in complex three-dimensional
  scenes.
\newblock {\em Image and Vision Computing}, 16(9):635--651, 1998.

\bibitem{jordan2012dpmeans}
B.~Kulis and M.~I. Jordan.
\newblock Revisiting k-means: New algorithms via {Bayesian} nonparametrics.
\newblock In {\em ICML}, 2012.

\bibitem{land1960automatic}
A.~H. Land and A.~G. Doig.
\newblock An automatic method of solving discrete programming problems.
\newblock {\em Econometrica: Journal of the Econometric Society},
  497--520, 1960.

\bibitem{lawler1966branch}
E.~L. Lawler and D.~E. Wood.
\newblock Branch-and-bound methods: A survey.
\newblock {\em Operations research}, 14(4):699--719, 1966.

\bibitem{li20073d}
H.~Li and R.~Hartley.
\newblock The 3D-3D registration problem revisited.
\newblock In {\em ICCV}, 2007.

\bibitem{liu1996quality}
A.~Liu and B.~Joe.
\newblock Quality local refinement of tetrahedral meshes based on
  8-subtetrahedron subdivision.
\newblock {\em AMS Math. Comp.}, 65(215):1183--1200, 1996.

\bibitem{magnusson2007scan}
M.~Magnusson, A.~Lilienthal, and T.~Duckett.
\newblock Scan registration for autonomous mining vehicles using 3D-NDT.
\newblock {\em Journal of Field Robotics}, 24(10):803--827, 2007.

\bibitem{magnusson2009evaluation}
M.~Magnusson, A.~N{\"u}chter, C.~L{\"o}rken, A.~J. Lilienthal, and
  J.~Hertzberg.
\newblock Evaluation of 3D registration reliability and speed-a comparison of
  ICP and NDT.
\newblock In {\em ICRA}, 2009.

\bibitem{makadia2006fully}
A.~Makadia, A.~Patterson, and K.~Daniilidis.
\newblock Fully automatic registration of 3D point clouds.
\newblock In {\em CVPR}, 2006.

\bibitem{mitra2004estimating}
N.~J. Mitra, A.~Nguyen, and L.~Guibas.
\newblock Estimating surface normals in noisy point cloud data.
\newblock {\em IJCGA}, 14:261--276, 2004.


\bibitem{newcombe2011kinectfusion}
R.~A. Newcombe, A.~J. Davison, S.~Izadi, P.~Kohli, O.~Hilliges, J.~Shotton,
  D.~Molyneaux, S.~Hodges, D.~Kim, and A.~Fitzgibbon.
\newblock Kinectfusion: Real-time dense surface mapping and tracking.
\newblock In {\em ISMAR}, 2011.

\bibitem{parra2014fast}
A.~J. Parra~Bustos, T.-J. Chin, and D.~Suter.
\newblock Fast rotation search with stereographic projections for 3D
  registration.
\newblock In {\em CVPR}, 2014.

\bibitem{Pomerleau:2012}
F.~Pomerleau, M.~Liu, F.~Colas, and R.~Siegwart.
\newblock {Challenging data sets for point cloud registration algorithms}.
\newblock {\em IJRR}, 31(14):1705--1711, 2012.

\bibitem{rusinkiewicz2001efficient}
S.~Rusinkiewicz and M.~Levoy.
\newblock Efficient variants of the ICP algorithm.
\newblock In {\em 3-D Digital Imaging and Modeling}, 2001.

\bibitem{rusu2009fast}
R.~B. Rusu, N.~Blodow, and M.~Beetz.
\newblock Fast point feature histograms (FPFH) for 3D registration.
\newblock In {\em ICRA}, 2009.

\bibitem{salvi2007review}
J.~Salvi, C.~Matabosch, D.~Fofi, and J.~Forest.
\newblock A review of recent range image registration methods with accuracy
  evaluation.
\newblock {\em Image and Vision Computing}, 25(5):578--596, 2007.

\bibitem{straub2015rtmf}
J.~Straub, N.~Bhandari, J.~J. Leonard, and J.~W. Fisher~III.
\newblock Real-time Manhattan world rotation estimation in 3D.
\newblock In {\em IROS}, 2015.

\bibitem{straub2015dpvmf}
J.~Straub, T.~Campbell, J.~P. How, and J.~W. Fisher~III.
\newblock Small-variance nonparametric clustering on the hypersphere.
\newblock In {\em CVPR}, 2015.

\bibitem{straub2014mmf}
J.~Straub, G.~Rosman, O.~Freifeld, J.~J. Leonard, and J.~W. Fisher~III.
\newblock A Mixture of Manhattan Frames: Beyond the Manhattan World.
\newblock In {\em CVPR}, 2014.

\bibitem{Teh10_EML}
Y.~W. Teh.
\newblock {{D}}irichlet processes.
\newblock In {\em Encyclopedia of Machine Learning}. Springer, New York, 2010.

\bibitem{tsin2004correlation}
Y.~Tsin and T.~Kanade.
\newblock A correlation-based approach to robust point set registration.
\newblock In {\em ECCV}, 2004.

\bibitem{turk1994zippered}
G.~Turk and M.~Levoy.
\newblock Zippered polygon meshes from range images.
\newblock In {\em SIGGRAPH}, 1994.

\bibitem{weiss1994keeping}
G.~Weiss, C.~Wetzler, and E.~Von~Puttkamer.
\newblock Keeping track of position and orientation of moving indoor systems by
  correlation of range-finder scans.
\newblock In {\em IROS}, 1994.

\bibitem{Whelan14ijrr}
T.~Whelan, M.~Kaess, H.~Johannsson, M.~Fallon, J.~Leonard, and J.~McDonald.
\newblock Real-time large scale dense {RGB-D SLAM} with volumetric fusion.
\newblock {\em IJRR}, 2014.

\bibitem{yang2013go}
J.~Yang, H.~Li, and Y.~Jia.
\newblock Go-ICP: Solving 3D registration efficiently and globally optimally.
\newblock In {\em ICCV}, 2013.

\bibitem{StellaSoft}
R.~Webb.
\newblock Stella software.
\newblock \url{http://www.software3d.com/Stella.php} and
  \url{https://en.wikipedia.org/wiki/600-cell}.

\bibitem{meshlab}
Meshlab.
\newblock \url{http://meshlab.sourceforge.net/}.
\newblock Accessed: 2016-11-15.

\bibitem{libpcl}
Point cloud library.
\newblock \url{http://pointclouds.org/}.
\newblock Accessed: 2016-11-15.

\end{thebibliography}

}

\onecolumn
\begin{center}
{\Large\textbf{Supplement}}
\end{center}
\appendix
\section{Rotational Alignment Details}

\subsection{The matrix $\Xi_{kk'}$}
In the main text, we are given two unit vectors $\mu_{1k}$ and $\mu_{2k'}$ in $\mathbb{R}^3$.
We define $\Xi_{kk'} = \Xi(\mu_{1k}, \mu_{2k'})$, where
$\Xi(u, v)\in\mathbb{R}^{4\times 4}$ is defined by
 $u^T(q\circ v) = q^T\Xi(u, v) q$,
where $u = (u_i, u_j, u_k)$, $v=(v_i, v_j, v_k)$, and
$q = (q_i, q_j, q_k, q_r)$. By standard quaternion rotation formula, we have
\begin{align*}
  \begin{aligned}
   u^T(q\circ v)
  &=\left[\begin{matrix}u_i \\u_j \\ u_k\end{matrix}\right]^T
  \left[\begin{matrix}
      1-2q_j^2-2q_k^2 & 2(q_iq_j-q_kq_r) & 2(q_iq_k+q_jq_r)\\
      2(q_iq_j+q_kq_r) & 1-2q_i^2-2q_k^2 & 2(q_jq_k-q_iq_r)\\
      2(q_iq_k-q_jq_r) & 2(q_jq_k+q_iq_r) & 1-2q_i^2-2q_j^2
  \end{matrix}\right]
  \left[\begin{matrix}v_i \\v_j \\ v_k\end{matrix}\right] \\
  &=q_i^2(-2u_jv_j-2u_kv_k) +
  q_j^2(-2u_iv_i-2u_kv_k) +
  q_k^2(-2u_iv_i-2u_jv_j) \\
  &+ q_iq_j(2u_jv_i+2u_iv_j) + q_jq_k(2u_kv_j+2u_jv_k) + q_iq_k(2u_iv_k+2u_kv_i)\\
  &+q_iq_r(2u_kv_j-2u_jv_k)+ q_jq_r(2u_iv_k-2u_kv_i) +
  q_kq_r(2u_jv_i-2u_iv_j) + u^Tv 
\end{aligned}
\end{align*}
Rearranging the quadratic expression in $q$ into the form $q^TMq$,
we find the formula for $\Xi(u, v)$:
\begin{align*}
  \Xi(u, v) &=\scalebox{0.8}{\mbox{\ensuremath{\displaystyle 
  \left[\begin{matrix}
      u_iv_i-u_jv_j-u_kv_k& u_jv_i+u_iv_j & u_iv_k+u_kv_i & u_kv_j-u_jv_k\\
      u_jv_i+u_iv_j & u_jv_j-u_iv_i-u_kv_k & u_jv_k+u_kv_j & u_iv_k-u_kv_i\\
      u_iv_k+u_kv_i & u_jv_k+u_kv_j & u_kv_k-u_iv_i-u_jv_j & u_jv_i - u_iv_j\\
     u_kv_j-u_jv_k & u_iv_k-u_kv_i & u_jv_i - u_iv_j & u^Tv
  \end{matrix}\right]}}}
\end{align*}
\subsection{Quadratic upper bound on $f$}

First, for any $z \in [a, b]$ where $0\leq a \leq b$, we can express  $z^2$ as a convex combination of $a^2$ and $b^2$, i.e.
\begin{align}
z^2 = \lambda a^2 + (1-\lambda)b^2 \implies \lambda = \frac{z^2-a^2}{b^2-a^2}
\end{align}
Since $f(\sqrt{z}) = \frac{e^{\sqrt{z}} -
  e^{-\sqrt{z}}}{\sqrt{z}}$ for $z \geq 0$ is convex (this can be shown by taking the second derivative
and showing it is nonnegative), we have
\begin{align}
f(z) = f\left(\sqrt{z^2}\right)
&= f\left(\sqrt{\lambda a^2 + (1-\lambda)b^2}\right)\\ &\leq \lambda f\left(a\right) + (1-\lambda)f\left(b\right)\\
&= z^2\left(\frac{f(b)-f(a)}{b^2-a^2}\right)
    +\left(\frac{b^2f(a)-a^2f(b)}{b^2-a^2}\right).
\end{align}
In the main text, since we know $\ell_{kk'} \leq z_{kk'}(q) \leq u_{kk'}$ for any $q\in\mathcal{Q}$,
we can use the above upper bound formula with $a = \ell_{kk'}$ and $b = u_{kk'}$.

\subsection{Derivation of the $\gamma_N$ bound}
\begin{lemma*}\label{lem:skewSup}
  Let $\gamma_N$ be the minimum dot product between any two tetrahedral vertices 
  at refinement level $N$. Then
  \begin{align}
    \frac{2\gamma_{N-1}}{1+\gamma_{N-1}} \leq \gamma_N.
  \end{align}
\end{lemma*}
\begin{proof}
  Let the vertices of the projected tetrahedron be $q_i$, $i\in\{1, 2, 3, 4\}$. Let $\gamma = \min_{j \neq k} q_j^Tq_k$, $\Gamma = \max_{j \neq k} q_j^Tq_k$ and
define the vertex between $q_i$ and $q_j$ as $q_{ij} =
\frac{q_i+q_j}{\|q_i+q_j\|}$. 
Upon subdividing the tetrahedron, there are three different types of edge in the new
  smaller tetrahedra.
\begin{figure}
  \begin{tikzpicture}
    \node (L0) at (0,0) {
      \includegraphics[width=0.3\columnwidth]{{./figures/tetrahedron_0_cropped}.png}
    };
    \node[right = 0.5ex of L0] (L1) {
      \includegraphics[width=0.3\columnwidth]{{./figures/tetrahedron_1_cropped}.png}
    };
    \node[right = 0.5ex of L1] (L2) {
      \includegraphics[width=0.3\columnwidth]{{./figures/tetrahedron_2_cropped}.png}
    };
  \end{tikzpicture}
  \caption{The three subdivision patterns of a
  tetrahedron displayed in 3D. Colors designate different edge types.
\label{fig:tetrahedronSubdivision}}
\end{figure}

Refer to Fig.~\ref{fig:tetrahedronSubdivision} for a depiction of these three types.

The first type of edge (blue in Fig.~\ref{fig:tetrahedronSubdivision}) is a \emph{corner edge} from a vertex to an edge midpoint. The cosine angle
between the vertices created by a corner edge is
  \begin{align}
    q_i^T q_{ij} = \sqrt{\frac{1+q_i^Tq_j}{2}} \geq
    \sqrt{\frac{1+\gamma}{2}}\,.
  \end{align}

The second type of edge (orange in
Fig.~\ref{fig:tetrahedronSubdivision}) is a \emph{tie edge} from an edge midpoint to an edge midpoint along a face.
The cosine angle
between the vertices created by a tie edge is
\begin{align}
  q_{ij}^T q_{ik}
  &=
  \frac{1+q_i^Tq_k+q_i^Tq_j+q_j^Tq_k}{2\sqrt{1+q_i^Tq_j}\sqrt{1+q_i^Tq_k}}\geq
  \frac{1+q_i^Tq_k+q_i^Tq_j+\gamma}{2\sqrt{1+q_i^Tq_j}\sqrt{1+q_i^Tq_k}} >
  \frac{1+3\gamma}{2(1+\gamma)}\,.
\end{align}
To see the rightmost inequality, consider the minimization
\begin{align}
  \min_{x, y}\;\frac{1+\gamma+x+y}{2\sqrt{1+x}\sqrt{1+y}} \quad
  \mathrm{s.t. }\; \gamma \leq x, y \leq \Gamma\,.
\end{align}
The optimum solution is at $x = y = \gamma$, since
the function is symmetric and monotonic in $x, y$: 
\begin{align}
  \begin{aligned}
    \frac{d}{dx}\left(\frac{1+\gamma+x+y}{2\sqrt{1+x}\sqrt{1+y}}\right) 
  &=\frac{1}{4\sqrt{1+x}\sqrt{1+y}}\left(1-
\frac{\gamma+y}{(1+x)}\right)\\
&> \frac{1}{4\sqrt{1+x}\sqrt{1+y}}\left(1-
\frac{\gamma+\Gamma}{1+\gamma}\right) > 0 .
\end{aligned}
\end{align}
The final type of edge (green in Fig.~\ref{fig:tetrahedronSubdivision})
is a \emph{skew edge} from an edge midpoint to an edge midpoint through
the interior of the tetrahedron. The cosine angle between vertices
created by a skew edge is
\begin{align}
 q_{ij}^T q_{kl} &=
 \frac{q_i^Tq_k+q_i^Tq_l+q_j^Tq_k+q_j^Tq_l}{2\sqrt{1+q_i^Tq_j}\sqrt{1+q_k^Tq_l}}\,.
\end{align}
Note that we can choose any of three skew edges in our refinement.
Therefore, we can formulate bounding the skew edge dot product as a
process where ``nature'' creates three skew edges, and we select the
best one (i.e. the one of maximum dot product).  Thus, in the worst
case, nature solves the following problem: given a selection of a skew
edge, minimize its dot product such that the other two dot products are
lower (and thus nature forces us to pick that edge).  Let 
\begin{align}
  \begin{aligned}
    s_{1} = q_1^Tq_3+q_2^Tq_4 \quad p_1 = (q_1^Tq_3)(q_2^Tq_4)\\
    s_{2} = q_1^Tq_4+q_2^Tq_3 \quad p_2 = (q_1^Tq_4)(q_2^Tq_3)\\
    s_{3} = q_1^Tq_2+q_3^Tq_4 \quad p_3 = (q_1^Tq_2)(q_3^Tq_4)\,.
  \end{aligned}
\end{align}
Then without loss of generality, we assume the ordering
\begin{align}
  \frac{s_1+s_2}{2\sqrt{1+s_3+p_3}} \geq \frac{s_1+s_3}{2\sqrt{1+s_2+p_2}}
  \geq \frac{s_2+s_3}{2\sqrt{1+s_1+p_1}}\,.
\end{align}
Now since the function $f(x, y)=(1+x)(1+y)$ constrained by $x+y=c$, $x, y
\geq 0$, reaches its maximum at $x=y=\frac{c}{2}$, we can reduce all of the
fractions above until $1+s_i+p_i=(1+s_i/2)^2$, and therefore redefining $x_i
= s_i/2$, this problem is reduced to minimizing the maximum fraction of
\begin{align}
  \frac{x_1+x_2}{1+x_3} \geq \frac{x_1+x_3}{1+x_2} \geq
  \frac{x_2+x_3}{1+x_1}\,.
\end{align}
Note that while the ordering of the inequalities may switch, we can assume
without loss of generality that the above holds (since we can simply redefine
labels 1, 2, and 3 accordingly). Next, note that the first inequality above
implies that $x_2 \geq x_3$,
and the second inequality likewise implies that $x_1\geq x_2$.
Therefore, minimizing over $x_1$ and $x_2$ while keeping $x_3$ fixed yields
\begin{align}
 \frac{2x_3}{1+x_3} \,.
\end{align}
And finally, minimizing over $x_3 \in \left[\gamma, \Gamma\right]$ yields
\begin{align}\label{eq:skewSup}
  \max_{\text{skew edges}} q_{ij}^T q_{kl} \geq
  \frac{2\gamma}{1+\gamma}\,.
\end{align}
For the final result of the proof, note that
\begin{align}
  \sqrt{\frac{1+\gamma}{2}}\geq \frac{1+3\gamma}{2(1+\gamma)}\geq
  \frac{2\gamma}{1+\gamma} \quad \forall \gamma\in\left[0, 1\right]\,.
\end{align}
\end{proof}

\subsection{Proof of Theorem 1 (rotational convergence)}
\begin{theorem*}\label{thm:epsconvergenceS3}
  Suppose $\gamma_0 = 36^\circ$ is the initial maximum angle between vertices in
  the tetrahedra tessellation of $\SDdim{3}$, and let
  \begin{align}
N &\triangleq \max\left\{0,
  \ceil[\Big]{\log_2\frac{\gamma_0^{-1}-1}{\cos\left(\epsilon/2\right)^{-1}-1}} \right\}\,.
\label{eq:S3searchDepthSup}
  \end{align}
  Then at most $N$ refinements are required to 
  achieve a rotational tolerance of $\epsilon$ degrees, and \BB{} has complexity $O(\epsilon^{-6})$.
\end{theorem*}

\begin{proof}
  Using Lemma~\ref{lem:skewSup}, we know that the minimum dot product
  between any two vertices in a single cover element $\mathcal{Q}$ 
  at refinement level $N$ satisfies
  \begin{align}
   \gamma_N \geq \frac{2\gamma_{N-1}}{1+\gamma_{N-1}}.
  \end{align}
  This function is monotonically increasing (by taking the derivative and showing it is positive). 
  So we recursively apply the bound:
  \begin{align}
   \gamma_N \geq \frac{2\frac{2\gamma_{N-2}}{1+\gamma_{N-2}}}{1+\frac{2\gamma_{N-2}}{1+\gamma_{N-2}}}
 = \frac{4 \gamma_{N-2}}{1+3\gamma_{N-2}} \geq \dots \geq \frac{2^N\gamma_0}{1+\left(2^N-1\right)\gamma_0}.
  \end{align}
  If we require a rotational tolerance of $\epsilon$ degrees,
  we need that $2\cos^{-1}\gamma_N \leq \epsilon$ (noting 
  that the rotation angle between two quaternions is 2 times
  the angle between their vectors in $\mathbb{S}^3$).
  Therefore, we need
  \begin{align}
    \gamma_N \geq \cos\left(\epsilon/2\right).
  \end{align}
  Using our lower bound, this is satisfied if
  \begin{align}
    \frac{2^N\gamma_0}{1+\left(2^N-1\right)\gamma_0} &\geq \cos\left(\epsilon/2\right) \implies
    N \geq \log_2 \frac{ \gamma_0^{-1} - 1}{\cos\left(\epsilon/2\right)^{-1} - 1}.
  \end{align}
  Since $N$ must be a nonnegative integer, the formula in Eq.~\eqref{eq:S3searchDepthSup} follows.
  At search depth $M$, the BB algorithm will have examined at most $M$ tetrahedra, where
  \begin{align}
    M&= 600(1+8+8^2+\dots+8^N) = 600\frac{8^{N+1}-1}{7}
  \end{align}
  Using the formula for $N$ in Eq.~\eqref{eq:S3searchDepthSup} (and noting $8=2^3$), we have
  \begin{align}
    M &= O\left( \left(\frac{\gamma_0^{-1}-1}{\cos(\epsilon/2)^{-1}-1}\right)^3\right) = 
      O\left( \left(\frac{\cos\left(\epsilon/2\right)}{ 1- \cos\left(\epsilon/2\right)}\right)^3\right).
  \end{align}
  Finally, using the Taylor expansion of cosine,
\begin{align}
    M &=
      O\left( \left(\frac{1-\epsilon^2}{\epsilon^2}\right)^3\right) =
        O\left(\epsilon^{-6}\right).
  \end{align}
\end{proof}

\subsection{Derivation for the $\ell_{kk'}$ and $u_{kk'}$ optimization}
We need to show that maximizing $\mu^T(q\circ \nu)$ for $q\in\mathcal{Q}$ 
is equivalent to maximizing $\mu^Tv$ for $v = M\alpha$, $\alpha \geq 0$, 
$\alpha \in \mathbb{R}^4$, for some $M\in\mathbb{R}^{3\times 4}$.
The following lemma establishes this fact.
\begin{lemma*}\label{lem:qtomSup}
  Let $\mathcal{Q}$ be a projected tetrahedron cover element on $\mathbb{S}^3$
  with vertices $q_i, \, i=1, \dots, 4$, define $m\in\mathbb{R}^3$
  satisfying $\|m\|=1$ (i.e.~$m\in\mathbb{S}^2$),
  and let $\mathcal{M}$ be the set of vectors reached by rotating $m$ by $q\in\mathcal{Q}$, 
  \begin{align}
\mathcal{M} \triangleq \left\{x \in
    \mathbb{R}^3 : x = q\circ m, q\in \mathcal{Q}\right\}.
  \end{align} 
  Then $\mathcal{M}$ can be described as a combination of vectors in
  $\mathbb{R}^3$ via
  \begin{align}
    \mathcal{M} = \left\{x \in \mathbb{R}^3 : \|x\|=1, \, x = M\alpha, \,
    \alpha\in\mathbb{R}^4_+\right\}.
  \end{align}
where $m_i \triangleq q_i\circ m \in \mathbb{R}^3$, and $M
\triangleq \left[m_1 \cdots m_4\right] \in \mathbb{R}^{3\times 4}$.
\end{lemma*}
\begin{proof}
  In this proof, we make use of quaternion notation.
  If $q = xi+yj+zk+w$ is a quaternion, then its pure component is 
  $\overrightarrow{q} = xi+yj+zk$, its scalar component is $\widetilde{q} = w$,
  and conjugation is denoted $q^*$.

  To begin the proof, note that $q \in \mathcal{Q}$ implies that $q =
  Q\alpha$ for some $\alpha\in\mathbb{R}^4_+$, by definition.  Since
  $q\circ m$ is a rotation of a vector, it returns a pure quaternion;
  thus,
  \begin{align}
    \begin{aligned}
      q\circ m &= \overrightarrow{q\circ m} = \overrightarrow{\sum_{i,
      j}\alpha_i\alpha_j q_im q_j^*} = \sum_{i,
      j}\alpha_i\alpha_j \overrightarrow{q_im q_j^*}\\
      &=\sum_{i, j}\alpha_i\alpha_j \overrightarrow{q_im q_i^*q_i q_j^*}
      =\sum_{i, j}\alpha_i\alpha_j \overrightarrow{m_i q_iq_j^*}
    \end{aligned}
  \end{align}
  where $\alpha_i$ is the $i^\text{th}$ component of $\alpha$.
  Now note that $q_iq_j^*$ is the quaternion that rotates $m_j$ to $m_i$:
  \begin{align}
 (q_iq_j^*) \circ m_j = (q_iq_j^*)m_j(q_iq_j^*)^* = q_iq_j^*q_jmq_j^*q_jq_i^*
 =q_im q_i^*
 =m_i.
  \end{align}
  Therefore, the axis of rotation of $q_iq_j^*$ is the unit vector directed
  along $m_j\times m_i$, and the angle is $\theta_{ij}$. Since $m_j\times
  m_i = \sin\left(\theta_{ij}\right)\widehat{m_j\times m_i}$, we have that 
  \begin{align} 
    q_iq_j^* = \left(m_j\times m_i
  \frac{\sin\left(\theta_{ij}/2\right)}{\sin\theta_{ij}}\right)^T\left[\begin{matrix}i\\j\\k\end{matrix}\right]
    + \cos{\frac{\theta_{ij}}{2}}w \,.
    \end{align}
    Using this expansion along with the identity $\overrightarrow{rs} = \tilde{r}\overrightarrow{s}
    + \tilde{s}\overrightarrow{r} +
    \overrightarrow{r}\times\overrightarrow{s}$,
  we have that
\begin{align}
    \begin{aligned}
      q\circ m &=\sum_{i, j}\alpha_i\alpha_j \overrightarrow{m_i q_iq_j^*}\\
      &=\sum_{i, j}\alpha_i\alpha_j \left( m_i
      \widetilde{q_iq_j^*} + m_i\times\overrightarrow{q_iq_j^*}\right)\\
      &=\sum_{i} \alpha_i^2 m_i  + \sum_{i \neq j}\alpha_i\alpha_j \left( m_i
      \widetilde{q_iq_j^*} + m_i\times\overrightarrow{q_iq_j^*}\right)\\
      &=\textstyle\sum_{i} \alpha_i^2 m_i  + \sum_{i < j}\alpha_i\alpha_j 
      \left( \left(m_i+m_j\right)
      \cos\left(\frac{\theta_{ij}}{2}\right)\right.\\
      &+\left.\tfrac{\sin\left(\theta_{ij}/2\right)}{\sin\theta_{ij}}
      \left(m_i\times\left(m_j\times
      m_i\right)+m_j\times\left(m_i\times
      m_j\right)\right)\right)
    \end{aligned}
  \end{align}
Now noting that  for any unit vectors $a, b\in\mathbb{R}^3$ with angle $\theta$ between them, we have 
\begin{align}
a\times(b\times a) = b - (\cos\theta) a
\end{align}
which can be derived from the triple product expansion identity $a\times (b\times c) = b(a\cdot c)-c(a\cdot b)$.
So applying this to  $m_i\times\left(m_j\times m_i\right)$ and $m_j\times\left(m_i\times m_j\right)$
\begin{align}
    \begin{aligned}
      q\circ m &=\sum_{i} \alpha_i^2 m_i  + \sum_{i < j}\alpha_i\alpha_j 
      \left( \left(m_i+m_j\right)
      \cos\left(\frac{\theta_{ij}}{2}\right)\right. \\
      &+\left. \frac{\sin\left(\theta_{ij}/2\right)}{\sin\theta_{ij}}
      \left(m_j - \cos\theta_{ij}m_i + m_i - \cos\theta_{ij}m_j\right)\right)
    \end{aligned}
  \end{align}
  and finally using the double angle formulas,
  \begin{align}
    \begin{aligned}
      q\circ m &=\sum_{i} \alpha_i^2 m_i  + \sum_{i < j}\alpha_i\alpha_j 
      \left( \left(m_i+m_j\right)
      \sec\left(\frac{\theta_{ij}}{2}\right)\right)
    \end{aligned}
  \end{align}
  combining, thus
\begin{align}
    \begin{aligned}
      q\circ m&=\sum_{i, j} \alpha_i\alpha_j m_i
      \sec\left(\frac{\theta_{ij}}{2}\right) 
    \end{aligned}
  \end{align}
  Since $\sec(\theta) \geq 0 \, \, \forall \theta \in \left(-\frac{\pi}{2},
  \frac{\pi}{2}\right)$, the coefficients are $\geq 0\, \, \forall \theta_{ij} \in
  \left(-\pi, \pi\right)$. Therefore, $q\circ m$ is a linear combination of the
  vectors $m_i$ with nonnegative coefficients.
\end{proof}

\section{Translational Alignment Derivations and Proofs}
Recall that we reuse notation in this section from the rotational section to
simplify the discourse and draw parallels to the rotational problem.
\subsection{Linear upper bound on $f$}
  For any $z \in \left[a, b\right]$ where $0\leq a \leq b$,
  we can express $z$ as a convex combination of $a$ and $b$, i.e.
\begin{align}
z = \lambda a + (1-\lambda) b \implies \lambda = \frac{z-a}{b-a}.
\end{align}
And, since $f(z) = e^z$ is convex,
  \begin{align}
  f(z) = f(\lambda a+(1-\lambda)b) &\leq\lambda f(a)+(1-\lambda)f(b)\\
&=  z\left(\frac{f(b)-f(a)}{b-a}\right)+\left(\frac{bf(a)-af(b)}{b-a}\right).
\end{align}
In the main text, since we know $\ell_{kk'} \leq z_{kk'}(q) \leq u_{kk'}$ for
any $q \in \mathcal{Q}$, we can use the above upper bound formula 
with $a = \ell_{kk'}$ and $b = u_{kk'}$.

\subsection{Proof of Theorem 2 (translational convergence)}
For translation, we have a similar result to Lemma~\ref{lem:skewSup},
but it is much simpler to show; the diagonal of each rectangular cell is
simply $1/2$ that of the previous refinement level, i.e.
\begin{align}
\frac{\gamma_{N-1}}{2} = \gamma_N = \Gamma_N = \frac{\Gamma_{N-1}}{2}.
\end{align}
\begin{theorem*}\label{thm:epsconvergenceR3}
  Suppose $\gamma_0$ is the initial diagonal of the translation cell
  in $\mathbb{R}^3$, and let
  \begin{align}\label{eq:nthmresultSup}
      N &\triangleq \max\left\{0, \ceil[\Big]{\log_2\frac{\gamma_0}{\epsilon}} \right\} \,.
  \end{align} 
  Then at most $N$ refinements are required to achieve 
a translational tolerance of $\epsilon$,
  and \BB{} has complexity $O(\epsilon^{-3})$. 
\end{theorem*}
\begin{proof}
  If $\gamma_0$ is the initial diagonal length, then
  $\gamma_N = 2^{-N}\gamma_0$.
  So to achieve a translational tolerance of
  $\epsilon$, we need that $\gamma_N \leq \epsilon$, meaning
  \begin{align}
    2^{-N}\gamma_0 &\leq \epsilon \implies
    N \geq \log_2 \frac{\gamma_0}{\epsilon}.
  \end{align}
  Since $N$ must be at least 0 and must be an integer, the formula in the theorem follows.   
   As the branching factor at each refinement is 8, the BB algorithm at level $N$ 
  will have examined at most $M$ cells, where
  \begin{align}
    M&= 1+8+8^2+\dots+8^N = \frac{8^{N+1}-1}{7} \,.
  \end{align}
  Substituting the result in Eq.~\eqref{eq:nthmresultSup} (and noting $8 = 2^3$), we have
  \begin{align}
    M &= O\left( \left(\frac{\gamma_0}{\epsilon}\right)^3\right) = 
        O\left( \epsilon^{-3}\right).
  \end{align}
  \end{proof}

\end{document}